\documentclass[journal]{IEEEtran}

\usepackage{algorithm}
\usepackage{algorithmic}
\usepackage[n,landau,notions,ff,mm]{cryptocode}
\usepackage{makecell}

\usepackage{newfloat}
\usepackage{listings}

\newfloat{listing}{tb}{lst}{}
\floatname{listing}{Listing}

\usepackage[misc]{ifsym}
\usepackage{graphicx} % Required for inserting images
\usepackage{breakurl}
\usepackage{subfigure}

\usepackage{float}
\usepackage{url}
\usepackage{xcolor}
\usepackage{balance}
\usepackage{xspace}
\usepackage{amsmath}
\usepackage{hyperref}
\usepackage{amsfonts}% 
\usepackage{multirow}
\usepackage{graphicx}
\usepackage{booktabs}
\usepackage{caption}
\usepackage{array}
\usepackage{enumitem}
\setlist[itemize]{left=0pt}
\setlist[enumerate]{left=0pt}

\usepackage{color, colortbl}
\usepackage[first=0,last=9]{lcg}

\usepackage{todonotes}
\setuptodonotes{inline, color=blue!30}

\definecolor{Gray}{gray}{0.9}
\definecolor{LightCyan}{rgb}{0.88,1,1}

\makeatletter
\DeclareRobustCommand\onedot{\futurelet\@let@token\@onedot}
\def\@onedot{\ifx\@let@token.\else.\null\fi\xspace}
\def\etal{\emph{et al}\onedot}

\usepackage{graphicx}%
\usepackage{multirow}%
\usepackage{amsmath,amssymb,amsfonts}%
\usepackage{amsthm}%
\usepackage{mathrsfs}%
\usepackage{xcolor}%
\usepackage{textcomp}%
\usepackage{manyfoot}%
\usepackage{booktabs}%
\usepackage{listings}%
\theoremstyle{thmstyleone}%
\theoremstyle{thmstyletwo}%

\theoremstyle{thmstylethree}%

\usepackage{pifont}% http://ctan.org/pkg/pifont
\newcommand{\cmark}{\ding{51}}%
\newcommand{\xmark}{\ding{55}}%

\usepackage{acro}
\acsetup{single}

\DeclareAcronym{FL}{
  short = FL,
  long  = Federated Learning,
}
\newcommand{\FL}{\ac{FL}\xspace}

\DeclareAcronym{ZKP}{
  short = ZKP,
  long  = Zero-knowledge proof,
}
\newcommand{\ZKP}{\ac{ZKP}\xspace}
\newcommand{\ZKPs}{\acp{ZKP}\xspace}

\newcommand{\FF}{\mathbb{F}}

\newcommand{\zksnark}{zk-SNARK\xspace}
\newcommand{\statement}{\textit{st}\xspace}
\newcommand{\Relationship}{\textit{R}\xspace}

\newcommand{\witness}{\ensuremath{\mathsf{wit}}\xspace}
\newcommand{\cm}{\mathsf{cm}}

\newcommand{\vk}{\mathsf{vk}}
\newcommand{\sparam}{\ensuremath{1^\lambda}\xspace}

\newcommand{\fun}[1]{\textsc{#1}}

\newcommand{\key}[1]{\mathsf{#1}}

\newcommand{\pk}{\key{pk}\xspace}

\newcommand{\True}{\key{True}\xspace}
\newcommand{\False}{\key{False}\xspace}

\newcommand{\set}[1]{\ensuremath{\{#1\}}\xspace}

\def \ifempty#1{\def\temp{#1} \ifx\temp\empty }
\newcommand{\zkFL}{\ensuremath{\xspace\texttt{zkFL}}\xspace}
\begin{document}

\title{\zkFL: Zero-Knowledge Proof-based Gradient Aggregation for Federated Learning}

\author{Zhipeng Wang, Nanqing Dong, Jiahao Sun, William Knottenbelt, and Yike Guo~\IEEEmembership{Fellow,~IEEE}
\thanks{This work was supported in part by FLock.io under the FLock Research Grant. (Corresponding author: Nanqing Dong.)}
\thanks{Z.~Wang and W.~Knottenbelt are with the Department of Computing, Imperial College London, London, SW7 2AZ, UK. (emails: zhipeng.wang20@imperial.ac.uk, w.knottenbelt@imperial.ac.uk)}
\thanks{N.~Dong is with the Shanghai Artificial Intelligence Laboratory, Shanghai 200232, China. (email: dongnanqing@pjlab.org.cn)}
\thanks{J.~Sun is with FLock.io, London, WC2H 9JQ, UK. (email: sun@flock.io)}
\thanks{Y.~Guo is with the Department of Computer Science and Engineering, Hong Kong University of Science and Technology, Hong Kong SAR, China; and also with the Data Science Institute, Imperial College London, London, SW7 2AZ, UK. (email:yikeguo@ust.hk)}
}

\markboth{Journal of IEEE Transactions on Big Data,, Vol. 00, No. 0, Month 2020}
{First A. Author \MakeLowercase{\textit{et al.}}: Bare Demo of IEEEtai.cls for IEEE Journals of IEEE Transactions on Big Data,}

\maketitle

\begin{abstract}
Federated learning (FL) is a machine learning paradigm, which enables multiple and decentralized clients to collaboratively train a model under the orchestration of a central aggregator. FL can be a scalable machine learning solution in \emph{big data} scenarios. Traditional FL relies on the trust assumption of the central aggregator, which forms cohorts of clients honestly. However, a malicious aggregator, in reality, could abandon and replace the client's training models, or insert fake clients, to manipulate the final training results.
In this work, we introduce \zkFL, which leverages zero-knowledge proofs to tackle the issue of a malicious aggregator during the training model aggregation process. To guarantee the correct aggregation results, the aggregator provides a proof per round, demonstrating to the clients that the aggregator executes the intended behavior faithfully. To further reduce the verification cost of clients, we use blockchain to handle the proof in a zero-knowledge way, where miners (\emph{i.e.}, the participants validating and maintaining the blockchain data) can verify the proof without knowing the clients' local and aggregated models. The theoretical analysis and empirical results show that \zkFL achieves better security and privacy than traditional FL, without modifying the underlying FL network structure or heavily compromising the training speed.
\end{abstract}


\section{Introduction}\label{sec: intro}

\IEEEPARstart{F}{ederated} learning (FL) is a privacy-preserving machine learning paradigm that allows multiple clients to collaboratively train a global model without sharing their raw data~\cite{mcmahan2017communication, pejo2023quality, huang2022fedcke, jiang2022towards}. FL can be a scalable solution for machine learning in \emph{big data} scenarios~\cite{doku2019towards}, where large-scale data are generated and stored by multiple clients in different physical locations. In FL, each participant (\emph{i.e.}, client) performs local training on its own private dataset and communicates only the model updates to the central server (\emph{i.e.}, aggregator). This decentralized approach minimizes the need to transfer large volumes of data to the aggregator. The aggregator then aggregates the model updates and sends the updated global model back to the clients. This process repeats iteratively until the global model converges or a stopping criterion is fulfilled. During the cross-device FL process, participants need to place their trust in the aggregator to create cohorts of clients in a fair and unbiased manner. However, a potential vulnerability is that an actively malicious adversary with control over the aggregator could exploit this trust~\cite{kairouz2021advances}. For instance, adversaries could carry out a Sybil attack~\cite{cao2022mpaf} by simulating numerous fake client devices, and the adversary could also selectively favor previously compromised clients' model updates from the pool of available participants. These attacks have the potential to enable the adversary to manipulate the final training results in \FL, compromising the integrity of the learning process. Safeguarding against such threats is imperative to maintain the effectiveness and security of cross-device \FL.

\begin{figure*}
\centering
\includegraphics[width=1.8\columnwidth]{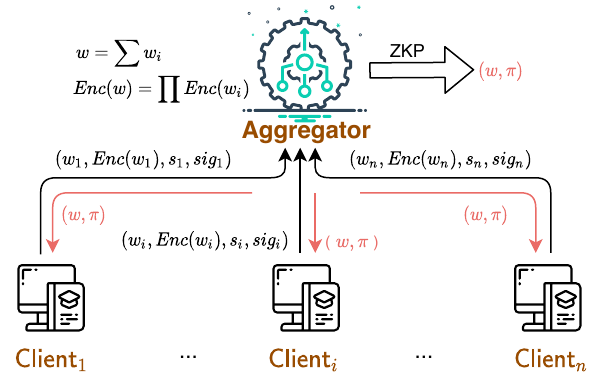}
\caption{Overview of \zkFL system. During each round, the clients send their local model update $w_i$, the encrypted model update $Enc(w_i)$, a chosen random number $s_i$, and their signature $sig_i$ to the aggregator.
The aggregator generates the aggregated model update $w$ and leverages zero-knowledge proofs (ZKPs) to generate a proof $\pi$, and the clients will verify the proof to ensure that the training model updates are correctly aggregated. \zkFL guarantees the integrity of the data computed by the aggregator, and enhances security and trust in the collaborative \FL setting.}
\label{fig:zkFL_without_blockchain}
\end{figure*}

\subsection{Contributions}
In this work, we present \zkFL (\emph{cf.}~Fig.~\ref{fig:zkFL_without_blockchain}), an innovative approach that integrates zero-knowledge proofs (ZKPs) into FL. Without changing the learning setup of the underlying FL method, this integration guarantees the integrity of aggregated data from the centralized aggregator.
% , all while preserving the inherent structure of the underlying FL network.
\ZKPs~\cite{groth2016size,sasson2014zerocash,DBLP:conf/sp/LycklamaBVKH23,liu2021zkcnn} are widely recognized cryptographic tools that enable secure and private computations while safeguarding the underlying data. In essence, ZKPs empower a prover to convince a verifier of a specific fact without revealing any information beyond that fact itself. Within the context of \zkFL, ZKPs play a pivotal role in addressing the challenge posed by a potentially malicious aggregator during the model aggregation process. To achieve accurate aggregation results, the aggregator must provide a proof for each round, demonstrating to the clients that it has faithfully executed the intended behavior for aggregating the model updates. By verifying these proofs, the clients can ensure the aggregator's actions are transparent and verifiable, instilling confidence that the aggregation process is conducted with utmost honesty.

\begin{figure*}[t]
\centering
\includegraphics[width=2\columnwidth]{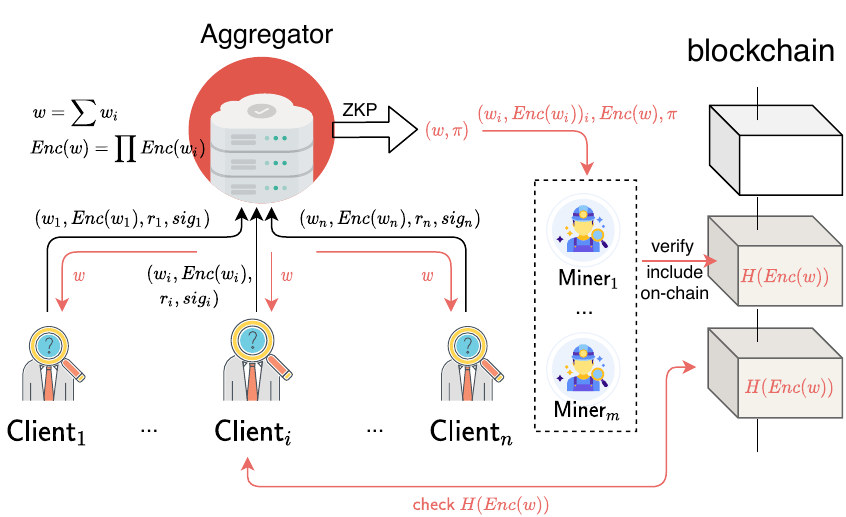}
\caption{Overview of \zkFL system with blockchain. 
The proof $\pi$ generated by the aggregator will be verified by the blockchain miners given the encrypted model updates $Enc(w_1), ..., Enc(w_n), Enc(w)$. This approach can reduce the computation cost of the clients, without compromising the privacy of the clients' models.}
\label{fig:zkFL_with_blockchain}
\end{figure*}

Furthermore, in order to minimize the verification burden on the clients, we propose a blockchain-based \zkFL solution to handle the proof in a zero-knowledge manner. As shown in Fig.~\ref{fig:zkFL_with_blockchain}, in this approach, the blockchain acts as a decentralized and trustless platform, allowing \emph{miners}, the nodes validating and maintaining the blockchain data~\cite{nakamoto2008bitcoin,bonneau2015sok, liu2020b4sdc}, to verify the authenticity of the \ZKP proof without compromising the confidentiality of the clients' models. By incorporating blockchain technology into our \zkFL system, we establish a robust and scalable framework for conducting zero-knowledge proof verification in a decentralized and transparent manner. This not only enhances the overall efficiency of the \zkFL system but also reinforces the confidentiality of the participants' data, making it a promising solution for secure and privacy-conscious cross-device \FL.

Our contributions can be summarized as follows: 
 \begin{itemize}
     \item We present \zkFL, an innovative \ZKP-based \FL system that can be integrated with existing FL methods. \zkFL empowers clients to independently verify proofs generated by the centralized aggregator, thereby ensuring the accuracy and validity of model aggregation results. \zkFL effectively addresses the threats posed by the malicious aggregators during the training model aggregation process, enhancing security and trust in the collaborative \FL setting.
     \item We integrate \zkFL with blockchain technology to minimize clients' computation costs for verification.  Leveraging the \emph{zero-knowledge} property of \ZKPs, our blockchain-based \zkFL significantly improves overall efficiency while preserving clients' model privacy,  thereby rendering it more scalable for \FL in big data environments.
     \item We present rigorous theoretical analysis on the security, privacy, and efficiency of \zkFL. We further evaluate these properties under benchmark \FL setups. The results of these experiments demonstrate the practical feasibility and effectiveness of \zkFL in real-world scenarios.
     % \item \towork{We conduct comprehensive and rigorous experiments to evaluate the performance of \zkFL on multiple well-known benchmark datasets.} The results of these experiments demonstrate the practical feasibility and effectiveness of \zkFL in real-world scenarios.
     
 \end{itemize}

\noindent\textbf{Paper Organization.} The remaining part of this paper is structured as follows. Section~\ref{sec: related-work} reviews related work in Zero-Knowledge Proofs (ZKPs) and Blockchain-based Federated Learning (FL). Section~\ref{sec: preliminary} outlines the preliminary concepts essential for our \zkFL and blockchain-based \zkFL frameworks. Our system and threat models are detailed in Section~\ref{sec: sys-and-threat-model}, followed by our methodology for developing \zkFL and blockchain-based \zkFL in Section~\ref{sec: security-analysis}. Sections~\ref{sec: security-analysis} and \ref{sec: implemenation} provide theoretical and empirical analyses of our constructions, respectively. Future directions are discussed in Section~\ref{sec: discussion}, and the paper concludes in Section~\ref{sec: conclusion}.

\section{Related Work} \label{sec: related-work}

\subsection{Zero-Knowledge Proofs} Zero-knowledge proofs (\ZKPs) have emerged as a revolutionary cryptographic concept that allows one party (the prover) to demonstrate the truth of a statement to another party (the verifier) without revealing any information beyond the statement's validity~\cite{groth2016size, sun2021survey}. One of the most notable applications of ZKPs is in privacy-preserving authentication, where a user can prove to an aggregator that they know a secret without revealing the secret itself. ZKPs have also been applied in other areas, including blockchains~\cite{sasson2014zerocash,wang2022zero}, machine learning~\cite{liu2021zkcnn,weng2023pvcnn,duan2023new} and FL. 

For instance, Burkhalter \etal introduce RoFL~\cite{DBLP:conf/sp/LycklamaBVKH23}, which is a secure \FL with ZKPs that enables the aggregator to enforce and verify constraints on client updates. Zhu \etal propose RiseFL~\cite{zhu2023robust} to ensure input data privacy and integrity from the \FL clients. Their approach employs a probabilistic integrity checking mechanism within ZKPs, combined with a hybrid commitment scheme, to enhance system performance effectively  Xing \etal propose  PZKP-FL~\cite{xing2023zero} which adopts \ZKPs to validate the computation process without disclosing the clients' local data in plaintext. 

\begin{table}[t]
    \centering
    \caption{Comparison of \zkFL and blockchain-based \zkFL with existing work.}
    \resizebox{\columnwidth}{!}{ 
    \begin{tabular}{l|ccc}
    \toprule
         Framework & Techniques & \makecell{Without\\ Trust in\\ Aggregator} & \makecell{Without \\On-Chain\\ Aggregation}\\  
         \toprule
         RoFL~\cite{DBLP:conf/sp/LycklamaBVKH23} &\ZKPs &\xmark &-\\
         RiseFL~\cite{zhu2023robust} &\ZKPs &\xmark &-\\ 
         PZKP-FL~\cite{xing2023zero} &\ZKPs + blockchain &\cmark &\xmark 
         \\
         stake-based FL~\cite{dong2024defending} & blockchain &\cmark &\xmark \vspace{1mm}\\
         
         \hline
         
         \zkFL &\ZKPs &- &\cmark\\
         blockchain-based \zkFL &\ZKPs + blockchain &\cmark &\cmark\\
    \bottomrule
    \end{tabular}
    }
        \label{tab:existing-work}
\end{table}

However, as shown in Table~\ref{tab:existing-work}, existing research of \ZKPs-based \FL designs~\cite{DBLP:conf/sp/LycklamaBVKH23, zhu2023robust}, has predominantly focused on using \ZKPs to address malicious client behaviors or enhance the client privacy, while assuming the existence of a ``honest-but-curious'' (\emph{i.e.}, semi-honest) aggregator. In this work, we break away from this assumption and leverage \ZKPs to ensure the honest aggregation of the centralized aggregator without requiring trust in the aggregator itself.

\subsection{Blockchain-based Federated Learning} Blockchain is a decentralized and immutable distributed ledger technology~\cite{huang2021survey,bonneau2015sok} that underpins cryptocurrencies such as Bitcoin~\cite{nakamoto2008bitcoin} and Ethereum~\cite{wood2014ethereum}. It provides a secure and transparent way to record and verify transactions across a network of nodes. Blockchain has been integrated with FL to tackle the security and privacy issues in existing FL~\cite{zhu2023blockchain,issa2023blockchain,qu2022blockchain,kim2019blockchained,weng2019deepchain}. Specifically, blockchain is used to store and manage the training model's updates and the associated metadata securely and transparently. Instead of relying solely on a centralized aggregator to manage the model updates, the blockchain enables a decentralized and distributed consensus mechanism among the participating clients. However, existing blockchain-based FL designs rely heavily on the on-chain computation. For example, in the design of PZKP-FL~\cite{xing2023zero}, a secure sum
protocol on blockchain is used to achieve the public verification of global aggregation. Dong \etal \cite{dong2024defending} propose a \FL framework with blockchain-based staking and voting scheme to mitigate the malicious behaviors from \FL clients, in which the aggregation process is performed on-chain with smart contract (\emph{i.e.}, self-executing programs that run on a blockchain network and are triggered by blockchain events). Such on-chain aggregation will cause expensive costs for FL networks with a large number of parameters.
As shown in Table~\ref{tab:existing-work}, we propose blockchain-based \zkFL, which can address the scalability issue of blockchain-based FL by using \ZKPs to remove the on-chain aggregation process.

\section{Preliminaries}\label{sec: preliminary}

In this section, we present the cryptographic building blocks for our \zkFL systems.

\subsection{Hash Functions}

In this work, we consider a cryptographic hash function \( H \) characterized by \( H: \{0,1\}^* \to \{0,1\}^\lambda \), capable of mapping inputs of arbitrary length to outputs of fixed length. The collision resistance of \( H \) is defined such that for any probabilistic polynomial-time (PPT) algorithm \(\mathcal A\), the probability that \(\mathcal{A}\) finds distinct inputs \(x\) and \(x'\) such that \(H(x) = H(x')\) is negligible. This property holds even when \( \mathcal{A} \) has knowledge of other hash outputs. We will employ this collision-resistant hash function \( H \) in the construction of our blockchain-based \zkFL. For detailed insights into hash functions, readers are referred to~\cite{rogaway2004cryptographic}.

\subsection{Commitments} 
A commitment scheme enables an entity to conceal a value while committing to it, with the ability to disclose the value at a later time if desired. The commitment scheme typically comprises two rounds: the committing round and the revealing round. In the committing round, the client commits to specific values while ensuring their confidentiality from others. The client retains the option to reveal the committed value in the subsequent revealing round. A commitment scheme includes two algorithms: 
  % \algoGroupTwo{commit}{verify}
  \begin{itemize}[noitemsep]
    \item $\cm \leftarrow \fun{Commit}(m, r)$ accepts a message $m$ and a secret randomness $r$ as inputs and returns the commitment $\cm$. 
    \item $0/1 \leftarrow \fun{Verify}(m, r, \cm)$ accepts a message $m$, a commitment $\cm$ and a decommitment value $r$ as inputs, and returns $1$ if the commitment is opened correctly and $0$ otherwise.
  \end{itemize}

In this paper, we leverage Pedersen commitments~\cite{pedersen1991non,aranha2022eclipse} to compute the clients' commitments/encryption on their local training model updates. Specifically, given the model update $w_i$,  a client will encrypt the update by computing $Enc(w_i) = g^{w_i}\cdot h^{s_i}$, where $g$ and $h$ are public parameters and $s_i$ is a random number generated by the client.

\subsection{Zero-Knowledge Proof} 
Zero-knowledge proof~\cite{kilian1992note,goldreich1994definitions,sun2021survey} is a cryptographic primitive that allows a prover to convince the verifier about
the correctness of some assertions without providing any meaningful information to the verifier.  A zero-knowledge proof of some statement satisfies the following three properties: 
\begin{itemize}
  \item \emph{Completeness:} If the statement $\statement$ is true, an honest verifier will always be convinced by an honest prover.
  \item \emph{Soundness:} For false statements, a prover cannot convince the verifier (even if
the prover deviates from the protocol).
  \item \emph{Zero-knowledge:} No verifier learns anything other than the fact that $\statement$ is valid if it is true. In other words, knowing the statement, but not the secret, is enough to construct a scenario in which the prover knows the secret.
\end{itemize}

A zero-knowledge Succinct Non-interactive ARgument of Knowledge
(\zksnark) is a ``succinct'' non-interactive zero-knowledge proof (NIZK) for arithmetic circuit satisfiability. The construction of \zksnark is based on a field $\FF$ and an arithmetic circuit $C$. We adopt the definition of \zksnark from~\cite{sasson2014zerocash}:
An arithmetic circuit satisfiability problem of a circuit $C: \FF^n\times\FF^h\rightarrow \FF^l$
is captured by the relationship $\Relationship_C: \set{(x,{\witness})\in \FF^n\times \FF^h: C(x,\witness) = 0^l}$, with 
the language $\mathcal L_C = \set{x \in \FF^n: \exists~\witness \in \FF^h~s.t.~C(x,\witness) = 0^l}$. A \zksnark for an arithmetic circuit satisfiability problem consists of the following algorithms: 

\begin{itemize}
    \item $(\pk, \vk) \leftarrow \fun{KeyGen}(\sparam, C)$: On input the security parameter $\sparam$ and the arithmetic circuit $C$, this algorithm outputs a proving key $\pk$ and a verification key $\vk$.

    \item $\pi\leftarrow \fun{Prove}(\pk,x,\witness)$: Given the proving key $\pk$ and $(x, \witness) \in \Relationship_C$, this algorithm outputs a proof $\pi$ for the statement $x \in \mathcal{L}_C$.

    \item $\True/\False \leftarrow\fun{Verify}(\vk, \pi, x)$: Taking the verification key $\vk$, the proof $\pi$, and the statement $x$ as input, this algorithm outputs $\True$ if $\pi$ is a valid proof for the statement $x \in \mathcal{L}_C$; otherwise, it outputs $\False$. 
\end{itemize}

In addition to the fundamental properties of \emph{correctness, soundness,} and \emph{zero-knowledge} inherent in \ZKPs, a \zksnark can exhibit additional essential characteristics. One such crucial aspect is \textit{succinctness}~\cite{sasson2014zerocash}, which demonstrates that the computational complexity of the $\fun{Verify}(\cdot)$ algorithm is linear with the size of $x$, denoted as $O_{\lambda}(|x|)$\footnote{We adhere to the notation introduced in~\cite{sasson2014zerocash}: $O_{\lambda}(\cdot)$ conceals a fixed polynomial factor in $\lambda$.}. Furthermore, the proof $\pi$ generated by an honest prover maintains a constant size in relation to $x$, denoted as $O_{\lambda}(1)$.

\section{System and Threat Models} \label{sec: sys-and-threat-model}
In this section, we outline our system model, threat model, and system goals.
\subsection{System Model}

\begin{itemize}
    \item \textbf{Clients}: In the context of \FL, clients represent individual devices, such as smartphones, tablets, or computers, each possessing its local dataset. These datasets remain secure and never leave the clients' devices. Instead, the clients independently train their machine learning models on their local data and communicate only the model updates to the central aggregator. 
    \item \textbf{Aggregator}: The aggregator acts as a central entity responsible for aggregating these model updates from multiple clients and computing a global model. This global model is then sent back to the clients, ensuring that each client benefits from the collective knowledge of the entire network while preserving data privacy and security.
\end{itemize}

\subsection{Threat Model}
We consider a malicious aggregator that can choose not to honestly aggregate the local model updates from clients. The malicious aggregator can deviate from the protocol by:
\begin{itemize}
    \item \emph{Abandoning} the updates generated from one or several honest clients.
\item Creating fake model updates to \emph{replace} the updates generated from honest clients, 

\item  \emph{Inserting} fake model updates to the updates generated from honest clients. 
\end{itemize}

We would like to remark that the scope of this work centers around the malicious aggregator rather than the honest-but-curious one, with a specific emphasis on ensuring aggregation integrity. We also acknowledge the potential for the aggregator to carry out model inversion attacks on the clients, a topic we intend to delve into in a future study. 

\subsection{System Goals}

\begin{itemize}
    \item \textbf{Security}: The aggregator cannot abandon or replace the local model updates generated from honest clients, nor insert any fake model updates into the final aggregated model update. Otherwise, the clients will detect the malicious behaviors of the aggregator and halt the \FL training process.
    \item \textbf{Privacy}: Only the participants (\emph{i.e.}, the aggregator and clients) in the \FL system can know the aggregated model updates during each round. 

\end{itemize}

% \newpage

\section{Methodology} \label{sec: zkFL}

\subsection{\zkFL}
As shown in Fig.~\ref{fig:zkFL_without_blockchain}, our \zkFL system works as follows:

\begin{enumerate}
    \item  \textbf{Setup:} $N$ clients and one aggregator generate their private/public key pairs and set up communication channels. Each client knows the public keys of the other $n-1$ clients, and this setup can be achieved by using a public key infrastructure (PKI).

\item \textbf{Local Training, Encrypting, and Signing:} During each round, the $n$ clients train their models locally to compute the local model updates $w_1, w_2, …, w_n$. Each client encrypts their update $Enc(w_i) = g^{w_i}\cdot h^{s_i}$ using Pedersen commitment, where $g$ and $h$ are public parameters and $s_i$ is a random number generated by the client. The client signs the encrypted updates with their private key to generate a signature $sig_i$. The client then sends the tuple of local model update, the randomly generated number, encrypted local model update, and signature $(w_i, s_i, Enc(w_i), sig_i)$ to the aggregator.

\item \textbf{Global Aggregation and \ZKP Generation:} The aggregator aggregates the received local model updates $w_1, w_2, …, w_n$ to generate the aggregated global model update $w = \sum_{i=1}^{n} {w_i}$. The aggregator also computes the aggregated value of the encrypted global model update $Enc(w) = \prod_{i=1}^n Enc(w_i)$ and signs it with its private key to generate the signature $sig$. The aggregator then leverages \zksnark to issue a proof $\pi$ for the following statement and witness: 
\begin{equation*}
    \begin{cases}
      & \text{$statement = (Enc(w_1),sig_1, Enc(w_2), sig_2, $}\\
      &~~~~~~~~~~~~~~~~~~~~~~~~~\text{$..., Enc(w_n), sig_n, Enc(w))$}\\
     & \text{$witness = (w_1, s_1, w_2, s_2,..., w_n, s_n, w)$}
    \end{cases}       
\end{equation*}
where the corresponding circuit $C(statement, witness)$ outputs $0$ if and only if:

\begin{equation*}
    \begin{cases}
      $(i)$ & \text{$\forall 1\leq i \leq n, Enc(w_i) = g^{w_i} \cdot h^{s_i}$}\\
      $(ii)$ & \text{$w = \sum_{i=1}^{n} {w_i}$}\\
      $(iii)$ & \text{$sig_i$ is signed by the client $i$}
    \end{cases}       
\end{equation*}

% (1) $Enc(w_i) = g^{w_i} \cdot h^{s_i}$, and (2) $w = \sum_{i=1}^{n} {w_i}$ and (3) $sig_i$ is signed by the client $i$.

 \item \textbf{Global Model Transmission and Proof Broadcast:} The aggregator transfers the aggregated global model update $w$, its encryption $Enc(w)$ and the proof $\pi$ to the $n$ clients.

 % and broadcasts the proof $\pi$, and the encrypted global model update $Enc(w)$ to the miners over the P2P network.

\item \textbf{Verification:} Upon receiving the proof $\pi$ and the encrypted global model update $Enc(w)$ from the aggregator, the clients verify if $\pi$ is valid. When the verification is passed, the clients start their local training based on the aggregated global model update $w$.

% \item \textbf{On-Chain Reading:} When the next round starts, the newly selected $n$ clients read the blockchain to check if $Enc(w)$ is appended on-chain. When the check is valid, the clients start their local training based on the aggregated global model update $w$.

\end{enumerate}

\subsection{Blockchain-based \zkFL }

% To further reduce the computation cost of clients, we adopt blockchain into our \zkFL system, in which the blockchain \emph{miners} will verify the proof generated by the aggregator.  As shown in Figure~\ref{fig:zkFL_with_blockchain}, the blockchain-based \zkFL works as follows:

To decrease the computation burden on clients, we incorporate blockchain technology into our \zkFL system. In this approach, the verification of proofs generated by the aggregator is entrusted to blockchain \emph{miners}. Illustrated in Fig.~\ref{fig:zkFL_with_blockchain}, the blockchain-based \zkFL operates as follows:

\begin{enumerate}
    \item  \textbf{Setup:} $N$ clients and one aggregator generate their private/public key pairs, which correspond to their on-chain addresses.

\item \textbf{Selection:} For each round, $n$ clients are selected from the $N$ clients via Verifiable Random Functions~\cite{micali1999verifiable,bitansky2020verifiable}. The $n$ selected clients’ public keys are broadcasted to the underlying P2P network of the blockchain, which will be received and verified by the miners.

\item \textbf{Local Training, Encrypting, and Signing:} The $n$ selected clients train their models locally to compute the local model updates $w_1, w_2, …, w_n$. Each client encrypts their update $Enc(w_i) = g^{w_i}\cdot h^{s_i}$ using Pedersen commitment, where $g$ and $h$ are public parameters and $s_i$ is a random number generated by the client. The client signs the encrypted updates with their private key to generate a signature $sig_i$. The client then sends the tuple of local model update, the randomly generated number, encrypted local model update, and signature $(w_i, s_i, Enc(w_i), sig_i)$ to the aggregator.

\item \textbf{Global Aggregation and \ZKP Generation:} The aggregator aggregates the received local model updates $w_1, w_2, …, w_n$ to generate the aggregated global model update $w = \sum_{i=1}^{n} {w_i}$. The aggregator also computes the aggregated value of the encrypted global model update $Enc(w) = \prod_{i=1}^n Enc(w_i)$ and signs it with its private key to generate the signature $sig$. The aggregator then leverages \zksnark to issue a proof $\pi$ for the following statement and witness: 
\begin{equation*}
    \begin{cases}
      & \text{$statement = (Enc(w_1),sig_1, Enc(w_2), sig_2, $}\\
      &~~~~~~~~~~~~~~~~~~~~~~~~~\text{$..., Enc(w_n), sig_n, Enc(w))$}\\
     & \text{$witness = (w_1, s_1, w_2, s_2,..., w_n, s_n, w)$}
    \end{cases}       
\end{equation*}
% $statement = (Enc(w_1),sig_1, Enc(w_2), sig_2, ..., Enc(w_n), sig_n, Enc(w))$ ; $witness = (w_1, s_1, w_2, s_2,..., w_n, s_n, w)$ 

where the corresponding circuit $C(statement, witness)$ outputs $0$ if and only if:

\begin{equation*}
    \begin{cases}
      $(i)$ & \text{$\forall 1\leq i \leq n, Enc(w_i) = g^{w_i} \cdot h^{s_i}$}\\
      $(ii)$ & \text{$w = \sum_{i=1}^{n} {w_i}$}\\
      $(iii)$ & \text{$sig_i$ is signed by the client $i$}
    \end{cases}       
\end{equation*}

 \item \textbf{Global Model Transmission and Proof Broadcast:} The aggregator transfers the aggregated global model update $w$ and its encryption $Enc(w)$ to the $n$ clients, and broadcasts the proof $\pi$, and the encrypted global model update $Enc(w)$ to the miners over the P2P network.

\item \textbf{On-Chain Verification:} Upon receiving the proof $\pi$ and the encrypted global model update $Enc(w)$ from the aggregator, the miners verify $\pi$ and append the hash value of $H(Enc(w))$ to the blockchain if $\pi$  is valid.

\item \textbf{On-Chain Reading:} When the next round starts, the newly selected $n$ clients read the blockchain to check if $H(Enc(w))$ is appended on-chain. When the check is valid, the clients start their local training based on the aggregated global model update $w$.

\end{enumerate}

\section{Theoretical Analysis} \label{sec: security-analysis}
In the following, we provide analyses to show that our \zkFL and blockchain-based \zkFL systems can achieve the goals of security and privacy, while merely bringing an acceptable decrease in the FL training efficiency.

\subsection{Security Analysis} Our \zkFL and blockchain-based \zkFL can achieve the security goal through the following techniques:
\begin{itemize}
    \item The signatures $sig_i$ of each encrypted local update $Enc(w_i)$ ensure the local updates' integrity from the clients and prevent the aggregator from tampering with their content and the statement $statement = (Enc(w_1),sig_1,  ..., Enc(w_n),$\\ $sig_n, Enc(w))$.
    \item The \emph{completeness} and \emph{soundness} properties of the \ZKP proof $\pi$ play a critical role in safeguarding the aggregation process from adversarial manipulation by the aggregator. These properties ensure that the aggregator cannot deviate from the intended behaviors:
    \begin{itemize}
        \item If the aggregator abandons the model update generated from one client $j$, then the aggregated results will be $\hat{w} = \sum_{i=1, j\not=i}^{n} {w_i}$. In this case, the corresponding circuit $C(statement, witness)$ outputs $0$ and the proof $\pi$ will be invalid.
        
        \item If the aggregator replaces the model update generated from one client $j$ by $\hat{w_j}$, then the aggregated results will be $\hat{w} = \sum_{i=1, j\not=i}^{n} {w_i} + \hat{w_j}$. In this case, the corresponding circuit $C(statement, witness)$ outputs $0$ and the proof $\pi$ will be invalid. 

        \item If the aggregator inserts one fake model update generated $\hat{w_0}$, then the aggregated results will be $\hat{w} = \sum_{i=1}^{n} {w_i} + \hat{w_0}$. In this case, the corresponding circuit $C(statement, witness)$ outputs $0$ and the proof $\pi$ will be invalid. 
    \end{itemize}
\end{itemize}

Therefore, the proof $\pi$ will only be valid if the aggregator honestly conducts the local model updates aggregation to generate the global model update $w = \sum_{i=1}^{n} {w_i}$.

\subsection{Privacy Analysis}
In \zkFL, privacy is inherently ensured as only the aggregator and the participating clients are involved in the training and aggregation process, eliminating the need for external parties. As a result, only these authorized entities possess knowledge of the aggregated model updates at each round.

In the context of the blockchain-based \zkFL system, the blockchain miners receive encrypted the local model updates $Enc(w_i) \thickspace(1\leq i \leq n)$, the encrypted global model update $Enc(w)$, and the \ZKP proof $\pi$ from the aggregator. However, due to the \emph{zero-knowledge} property of \ZKP, the miners can only verify whether $Enc(w)$ is correctly executed or not, without gaining any access to information about the individual local model updates $w_i \thickspace (1\leq i \leq n)$ or the global model update $w$. Additionally, storing the encrypted data of $Enc(w)$ on the blockchain does not compromise the privacy of the global model update $w$. Our system maintains a robust level of privacy throughout the blockchain-based \zkFL process.

\subsection{Efficiency Analysis}\label{sec: Efficiency-Analysis}
In the following, we calculate the expected computation time of the aggregator and a client per round, to analyze the system efficiency of \zkFL and blockchain-based \zkFL.

In both \zkFL and blockchain-based \zkFL systems, the aggregator is responsible for aggregating the local model updates and generating the \ZKP proof. The expected computation time of the aggregator is:
$$\mathbb{E}_{aggregator} = \mathbb{E}(aggr) + \mathbb{E}(ZKP.gen)$$

In the \zkFL system, a client needs to train the local model, encrypt the local model update, and verify the \ZKP proof generated by the aggregator. The expected computation time of a client is:
$$\mathbb{E}_{client} = \mathbb{E}(train) +\mathbb{E}(enc) + \mathbb{E}(ZKP.ver)$$

In the blockchain-based \zkFL system, a client still needs to train the local model and encrypt the local model update. However, the blockchain miners will verify the \ZKP proof generated by the aggregator, and the clients only need to read the data on the blockchain. The expected computation time of a client is:
$$\mathbb{E}^{block}_{client} = \mathbb{E}(train) +\mathbb{E}(enc) + \mathcal{O}(1)$$

\section{Empirical Analysis} \label{sec: implemenation}
In this section, we quantify the overhead of \zkFL and show that it can be used to train practical \FL models.
\subsection{Experiment Setup}

\begin{figure*}[t]
% \centering
\subfigure[Training accuracy.]{
\includegraphics[width=0.48\columnwidth]{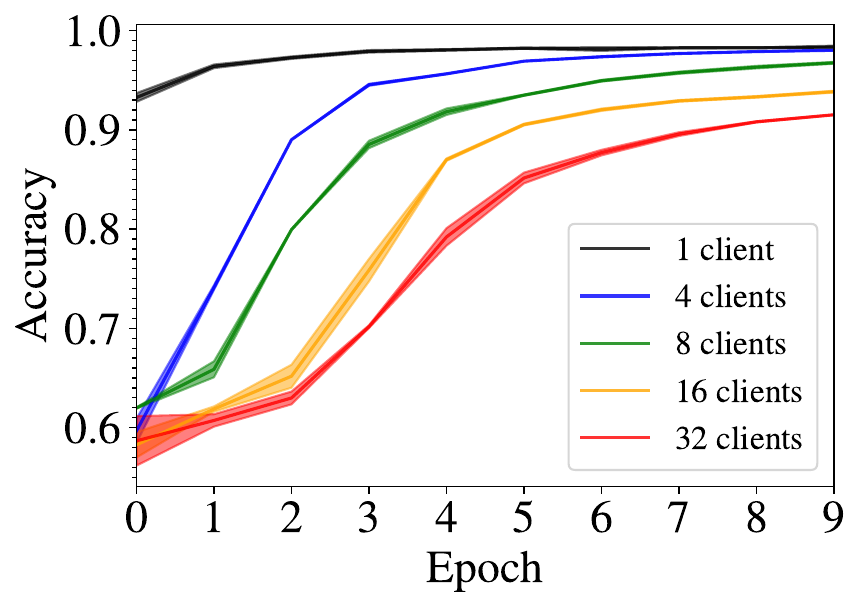}
\label{fig: resnet18_accuracy_time}
}%
\subfigure[Total training time.]{
\includegraphics[width=0.48\columnwidth]{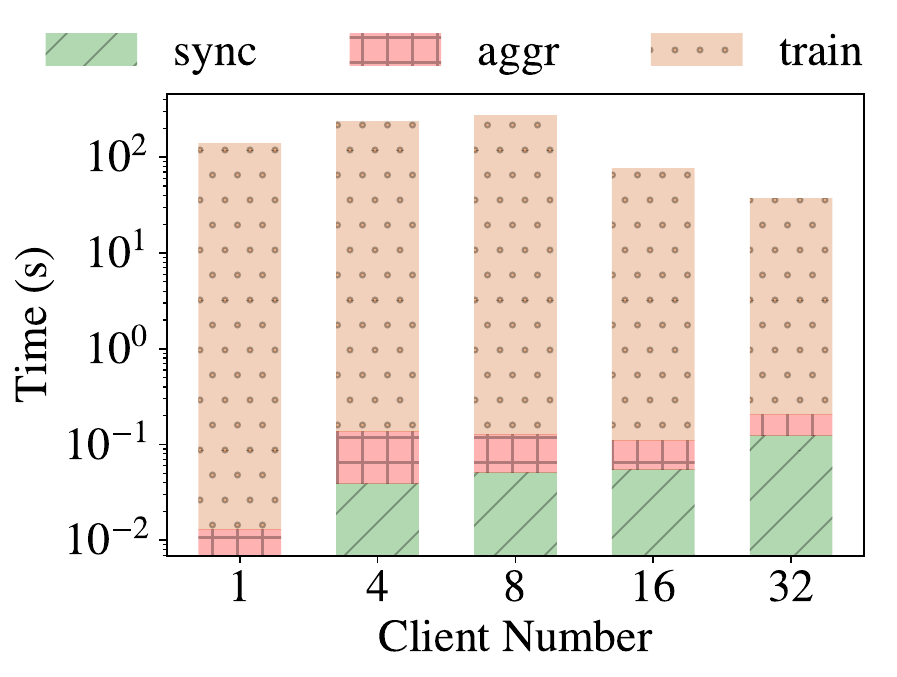}
\label{fig: resnet18_train_time}
}%
\subfigure[\zkFL encryption time.]{
\includegraphics[width=0.48\columnwidth]{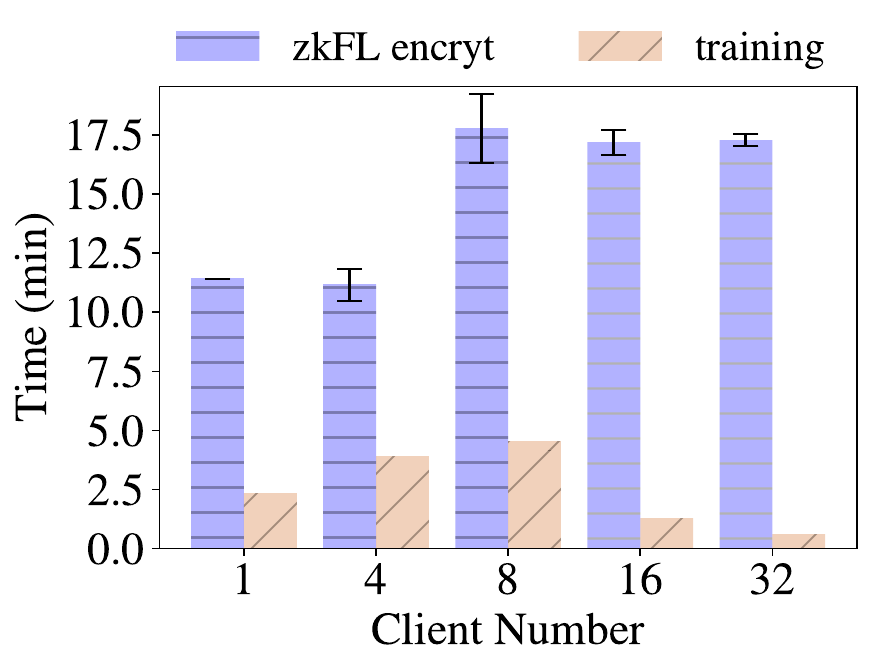}
\label{fig: resnet18_encryption_time}
}%
\subfigure[\zkFL aggregation time.]{
\includegraphics[width=0.48\columnwidth]{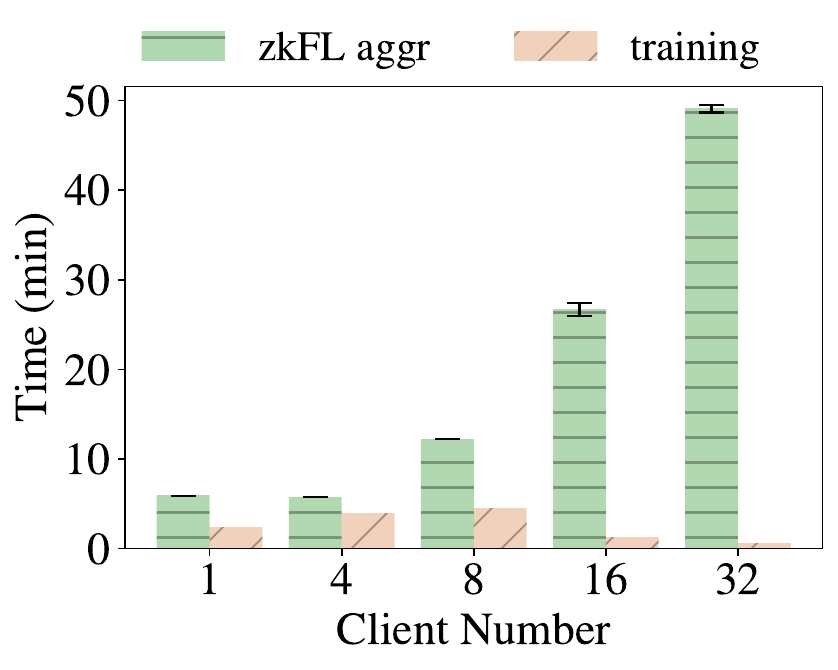}
\label{fig: resnet18_aggregation_time}
}%

\centering
\caption{Accuracy, training, encryption and aggregation time for Resnet18 on CIFAR-10 with \zkFL system.}
\label{fig:encryption}
\end{figure*}

\begin{figure*}[t]
\centering
\subfigure[Training accuracy.]{
\includegraphics[width=0.48\columnwidth]{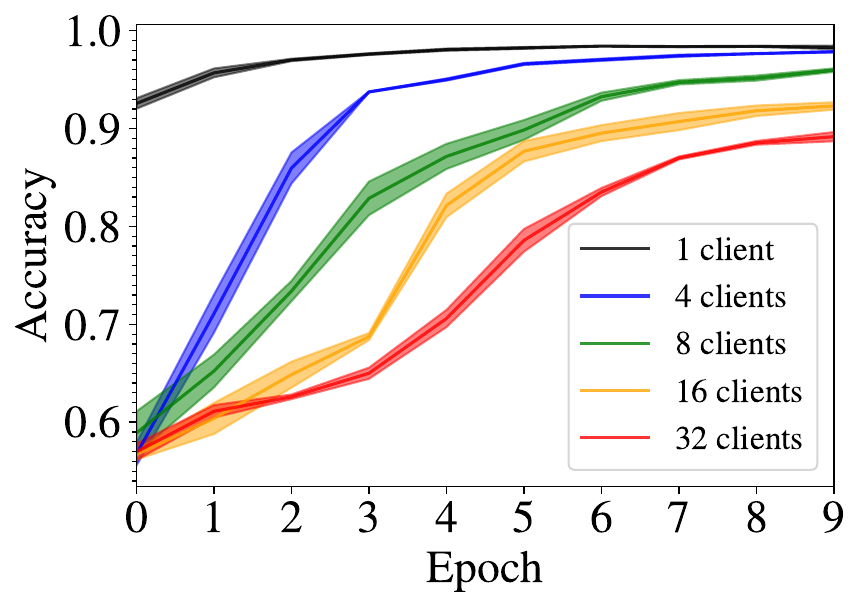}
\label{fig: resnet34_accuracy_time}
}%
\subfigure[Total training time.]{
\includegraphics[width=0.48\columnwidth]{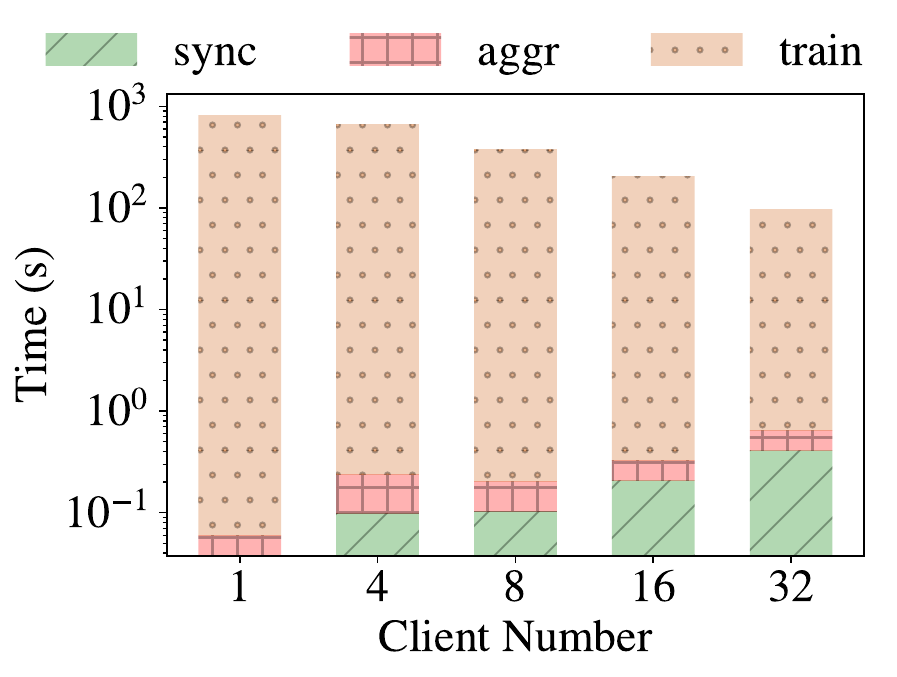}
\label{fig: resnet34_train_time}
}%
\hfill
\subfigure[\zkFL encryption time.]{
\includegraphics[width=0.48\columnwidth]{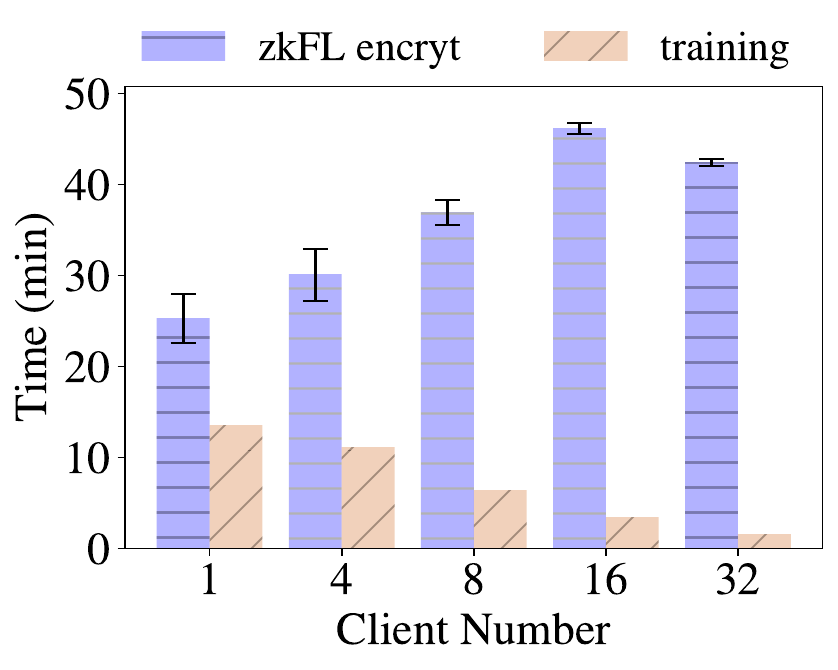}
\label{fig: resnet34_encryption_time}
}%
\subfigure[\zkFL aggregation time.]{
\includegraphics[width=0.48\columnwidth]{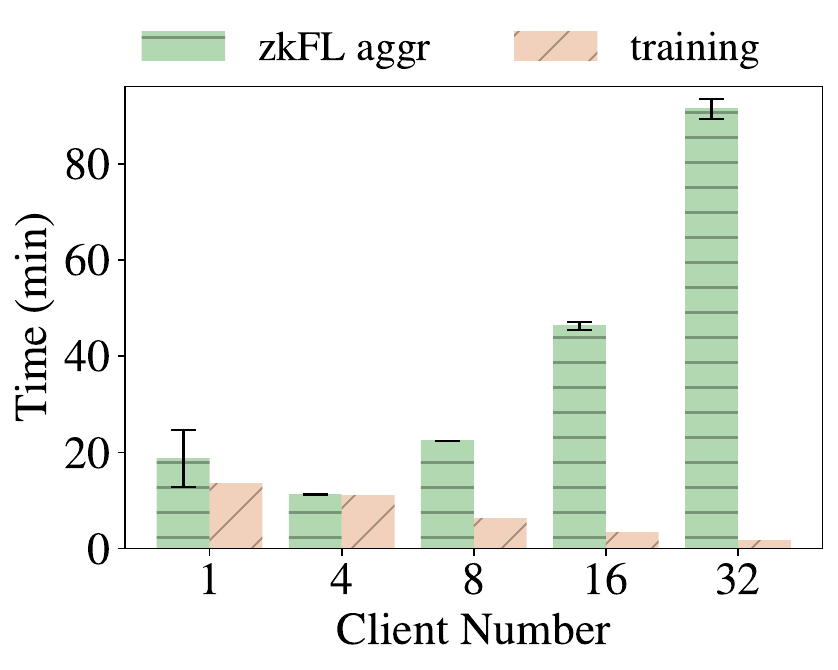}
\label{fig: resnet34_aggregation_time}
}%
\centering
\caption{Accuracy, training, encryption, and aggregation time for Resnet34 on CIFAR-10 with \zkFL system.}
\label{fig:encryption}
\end{figure*}

\begin{figure*}[t]
\centering
\subfigure[Training accuracy.]{
\includegraphics[width=0.48\columnwidth]{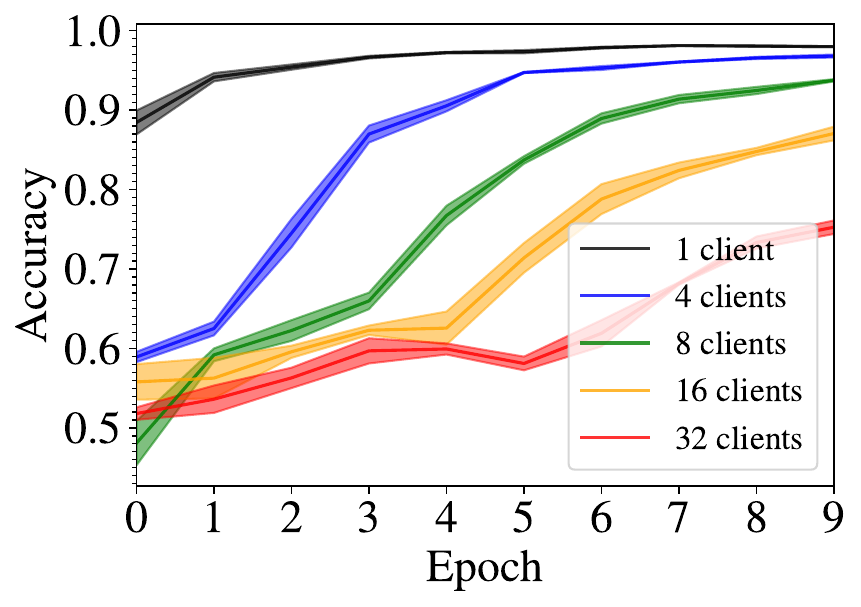}
\label{fig: resnet50_accuracy_time}
}%
\subfigure[Total training time.]{
\includegraphics[width=0.48\columnwidth]{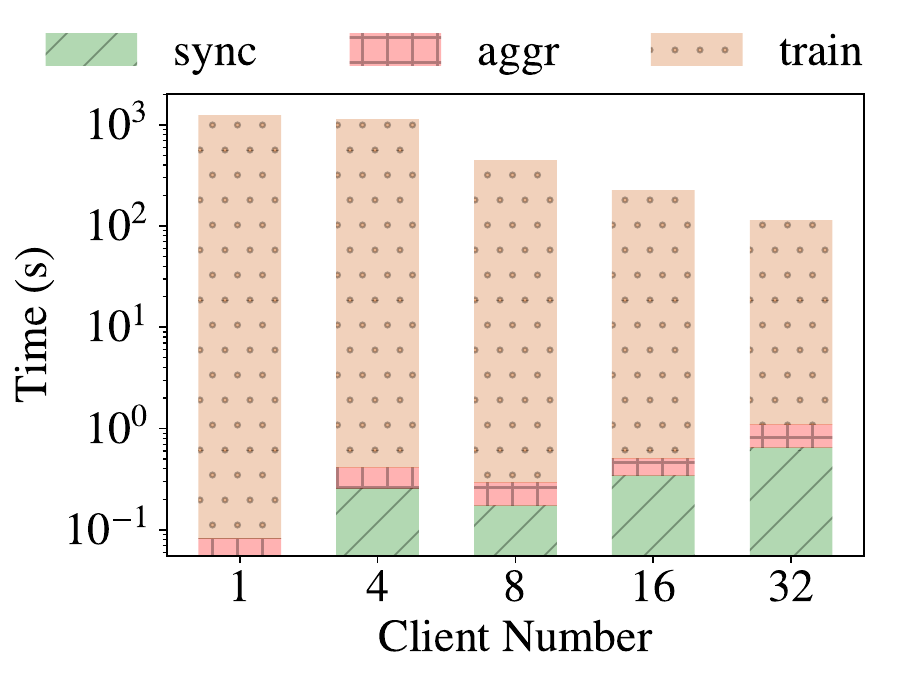}
\label{fig: resnet50_train_time}
}%
\hfill
\subfigure[\zkFL encryption time.]{
\includegraphics[width=0.48\columnwidth]{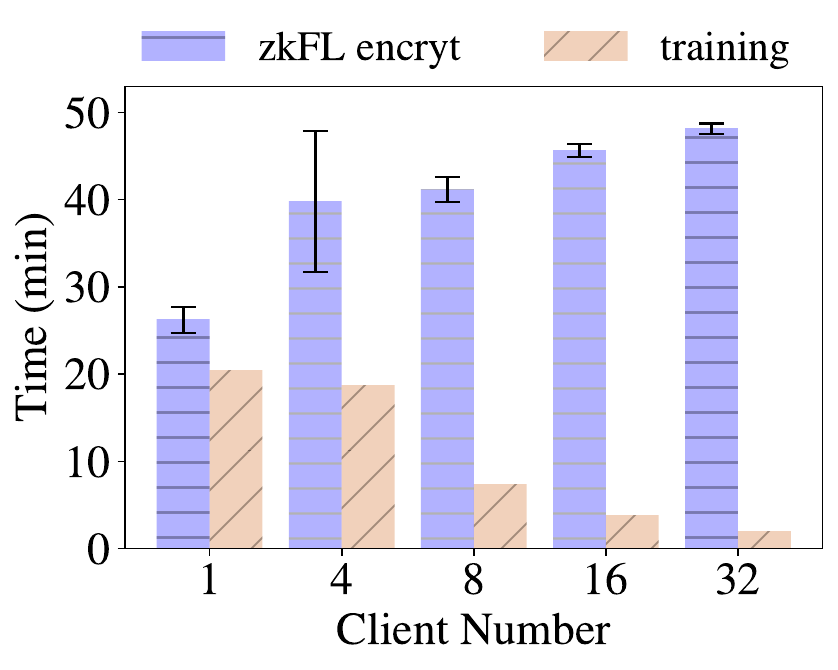}
\label{fig: resnet50_encryption_time}
}%
\subfigure[\zkFL aggregation time.]{
\includegraphics[width=0.48\columnwidth]{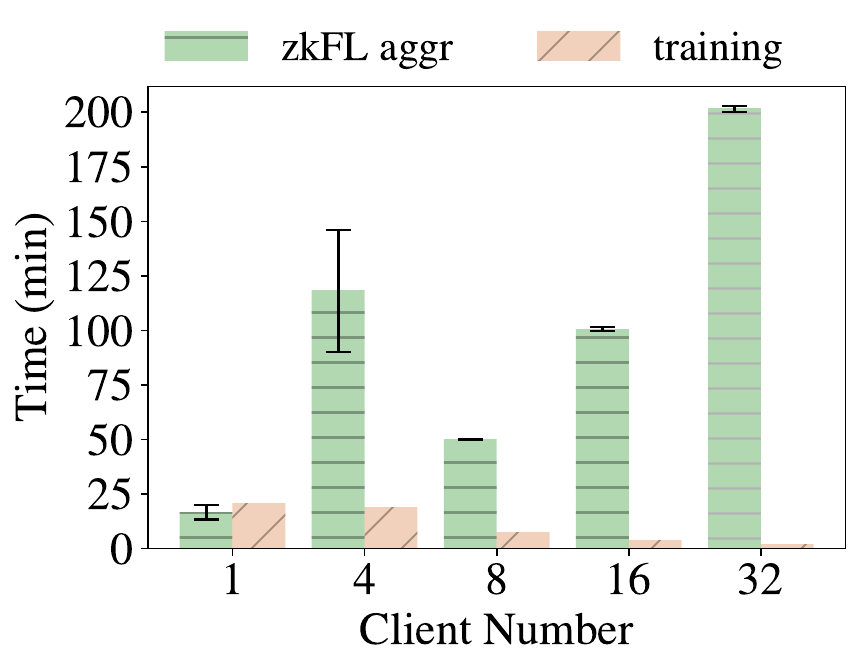}
\label{fig: resnet50_aggregation_time}
}%
\centering
\caption{Accuracy, training, encryption, and aggregation time for Resnet50 on CIFAR-10 with \zkFL system.}
\label{fig:encryption}
\end{figure*}

\begin{figure*}[t]
\centering
\subfigure[Training accuracy.]{
\includegraphics[width=0.48\columnwidth]{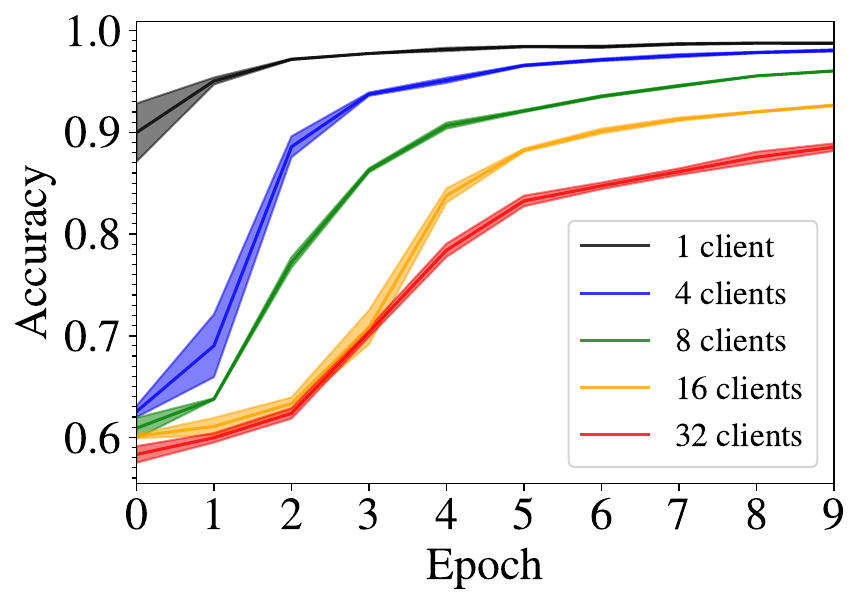}
\label{fig: densenet121_accuracy_time}
}%
\subfigure[Total training time.]{
\includegraphics[width=0.48\columnwidth]{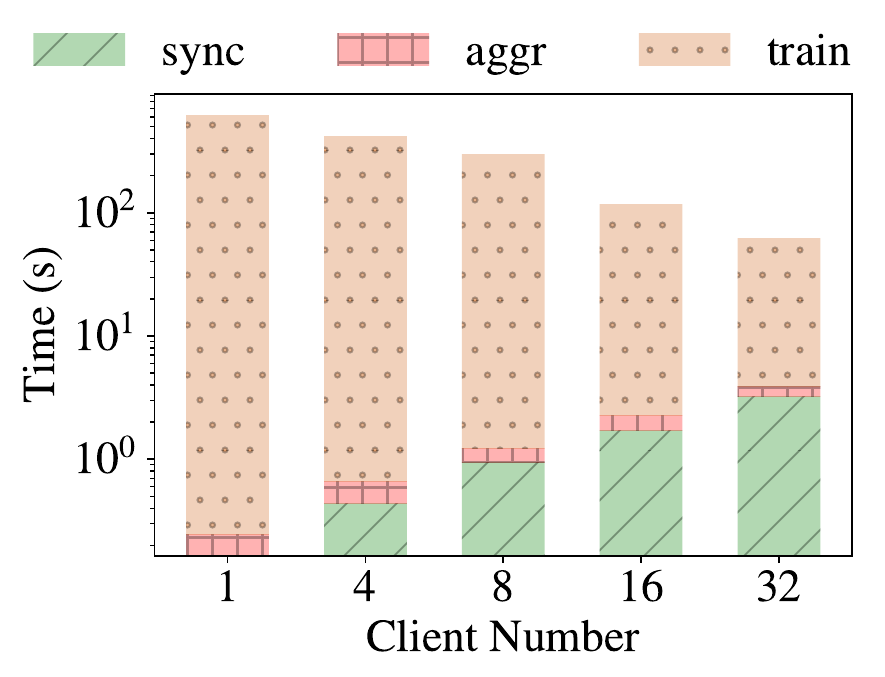}
\label{fig: densenet121_train_time}
}%
\hfill
\subfigure[\zkFL encryption time.]{
\includegraphics[width=0.48\columnwidth]{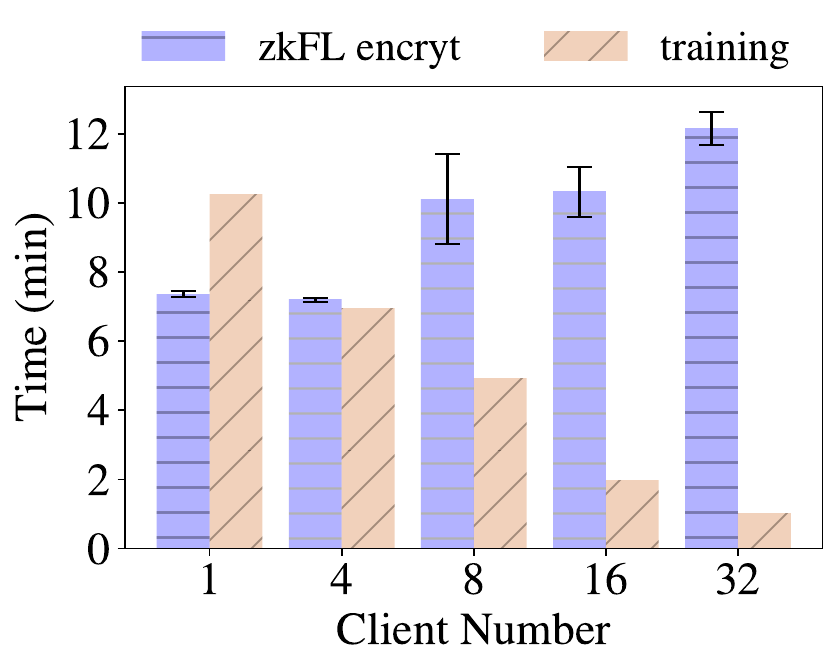}
\label{fig: densenet121_encryption_time}
}%
\subfigure[\zkFL aggregation time.]{
\includegraphics[width=0.48\columnwidth]{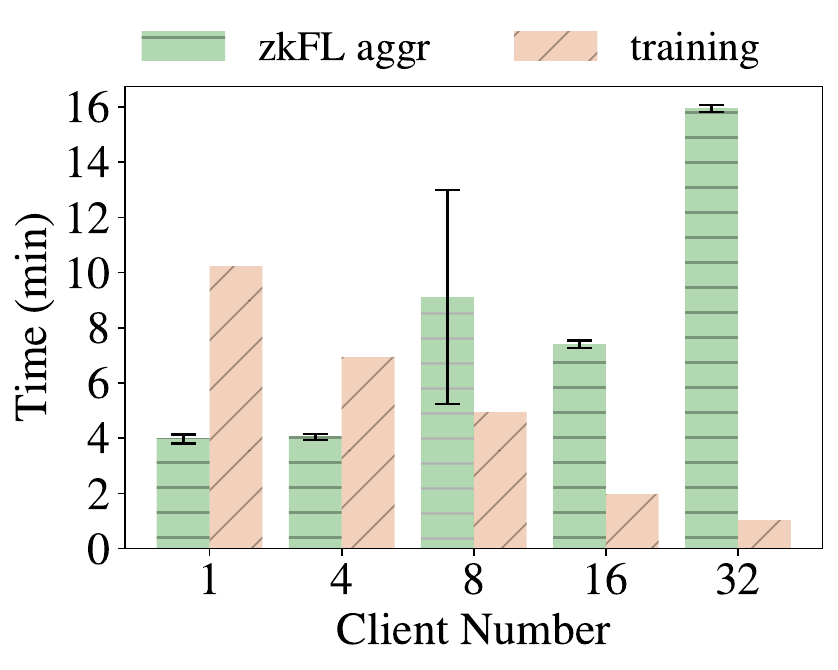}
\label{fig: densenet121_aggregation_time}
}%
\centering
\caption{Accuracy, training, encryption, and aggregation time for Densenet121 on CIFAR-10 with \zkFL system.}
\label{fig:encryption}
\end{figure*}

\begin{figure*}[t]
\centering
\subfigure[Training accuracy.]{
\includegraphics[width=0.48\columnwidth]{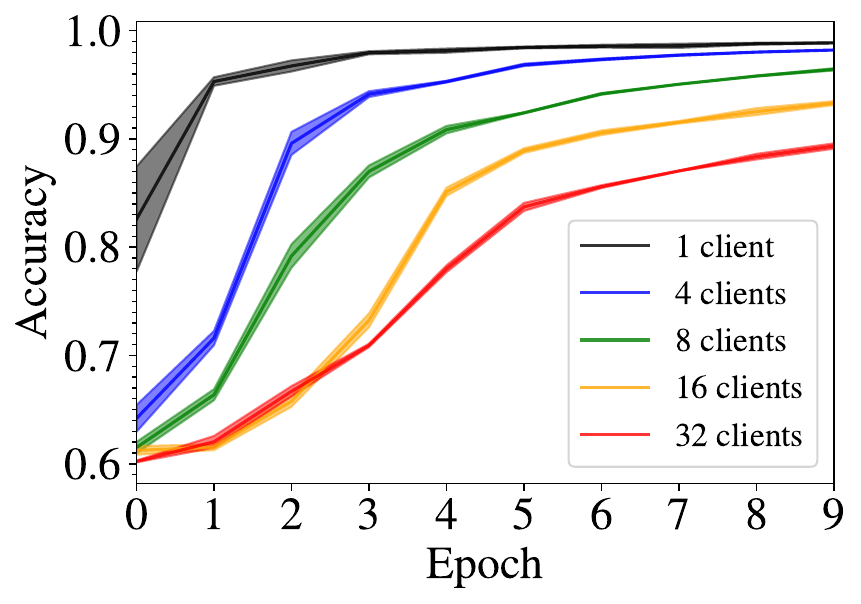}
\label{fig: densenet169_accuracy_time}
}%
\subfigure[Total training time.]{
\includegraphics[width=0.48\columnwidth]{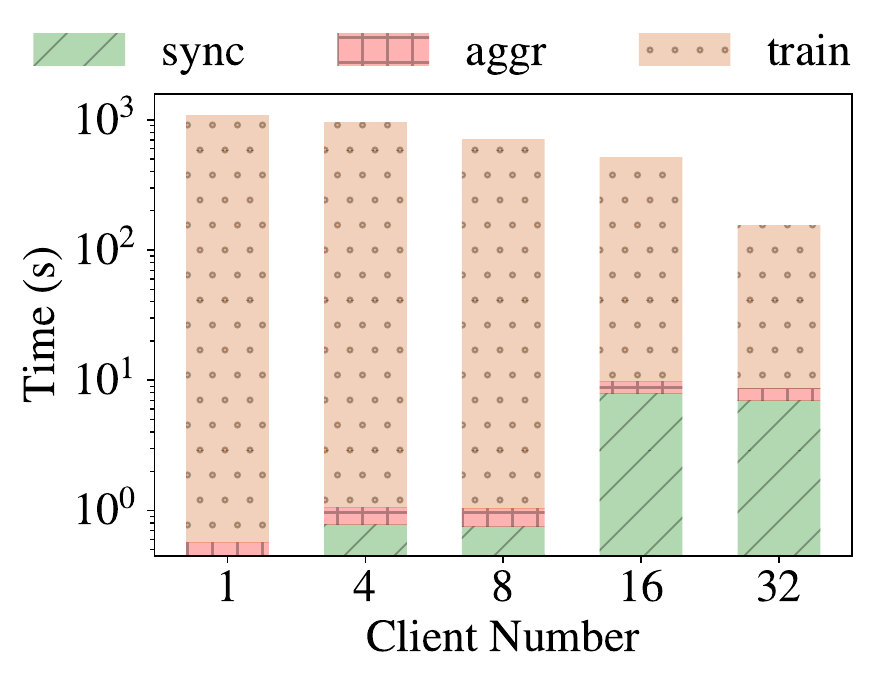}
\label{fig: densenet169_train_time}
}%
\hfill
\subfigure[\zkFL encryption time.]{
\includegraphics[width=0.48\columnwidth]{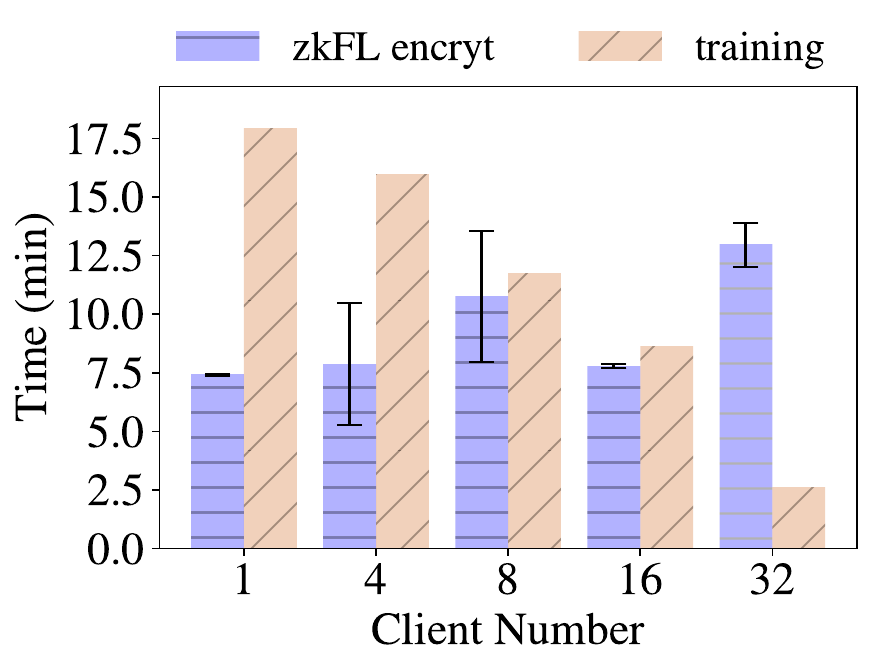}
\label{fig: densenet169_encryption_time}
}%
\subfigure[\zkFL aggregation time.]{
\includegraphics[width=0.48\columnwidth]{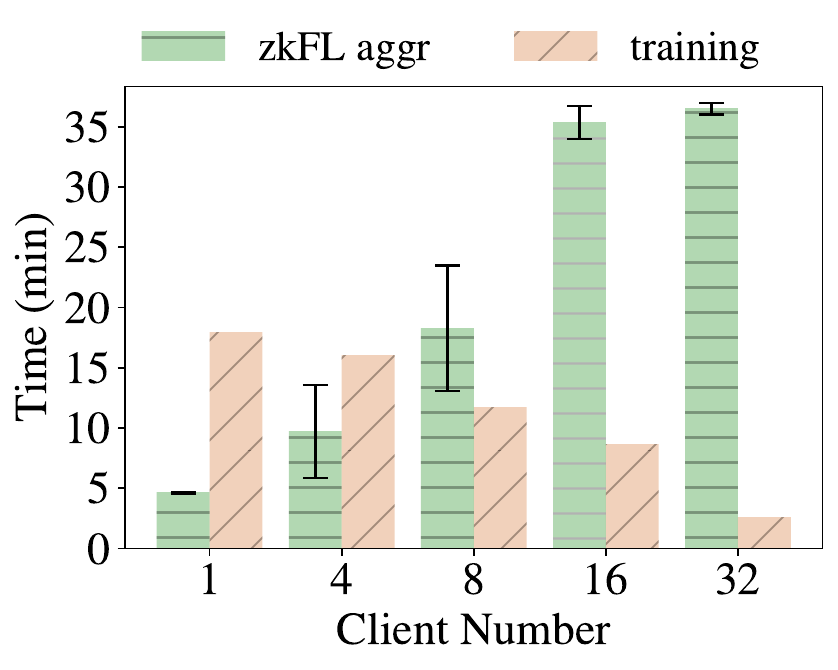}
\label{fig: densenet169_aggregation_time}
}%
\centering
\caption{Accuracy, training, encryption, and aggregation time for Densenet169 on CIFAR-10 with \zkFL system.}
\label{fig:encryption}
\end{figure*}

\begin{figure*}[t]
\centering
\subfigure[Training accuracy.]{
\includegraphics[width=0.48\columnwidth]{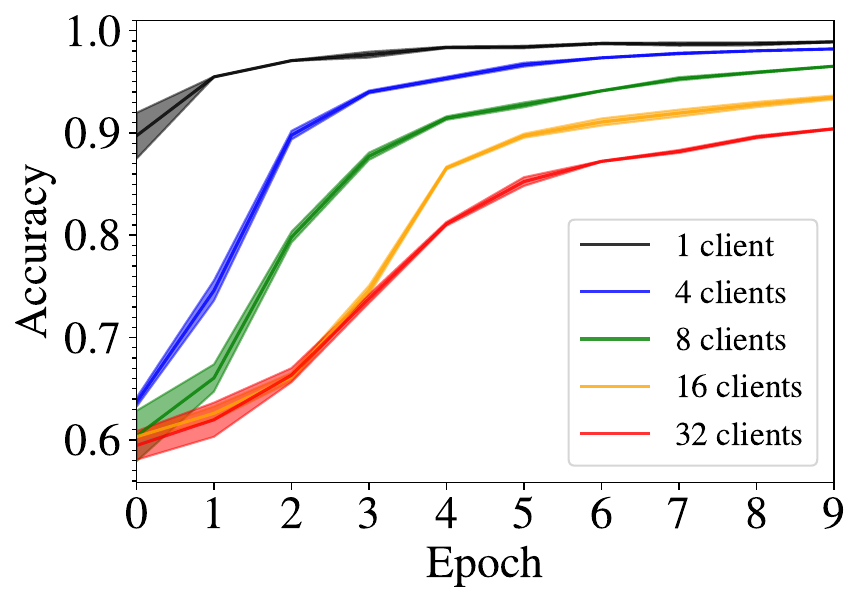}
\label{fig: densenet201_accuracy_time}
}%
\subfigure[Total training time.]{
\includegraphics[width=0.48\columnwidth]{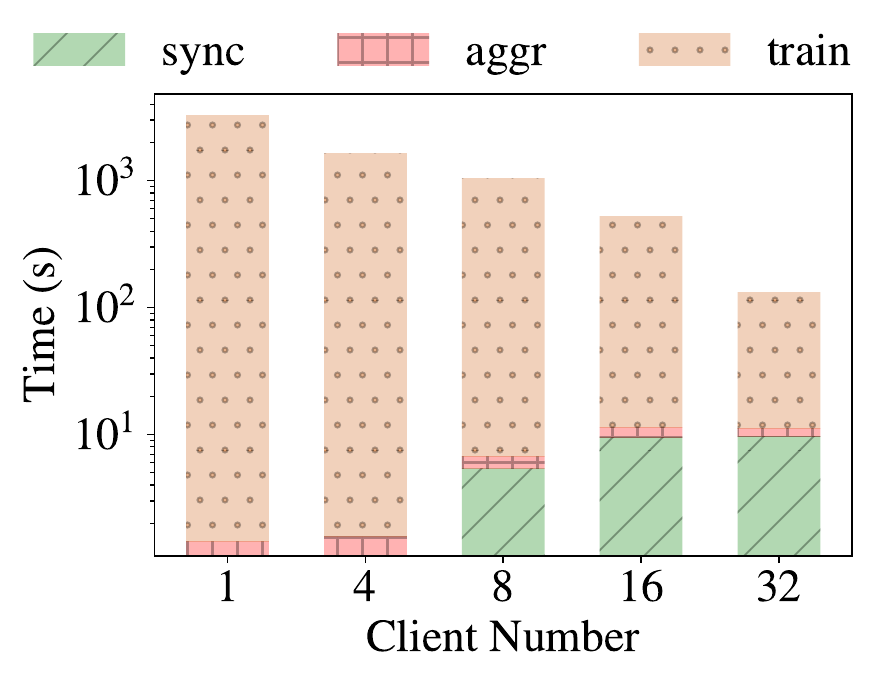}
\label{fig: densenet201_train_time}
}%
\hfill
\subfigure[\zkFL encryption time.]{
\includegraphics[width=0.48\columnwidth]{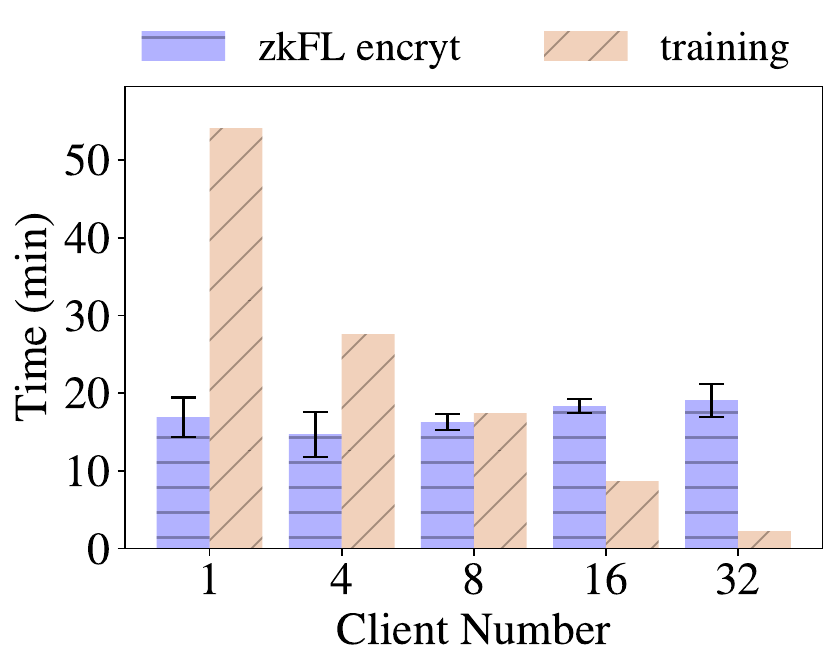}
\label{fig: densenet201_encryption_time}
}%
\subfigure[\zkFL aggregation time.]{
\includegraphics[width=0.48\columnwidth]{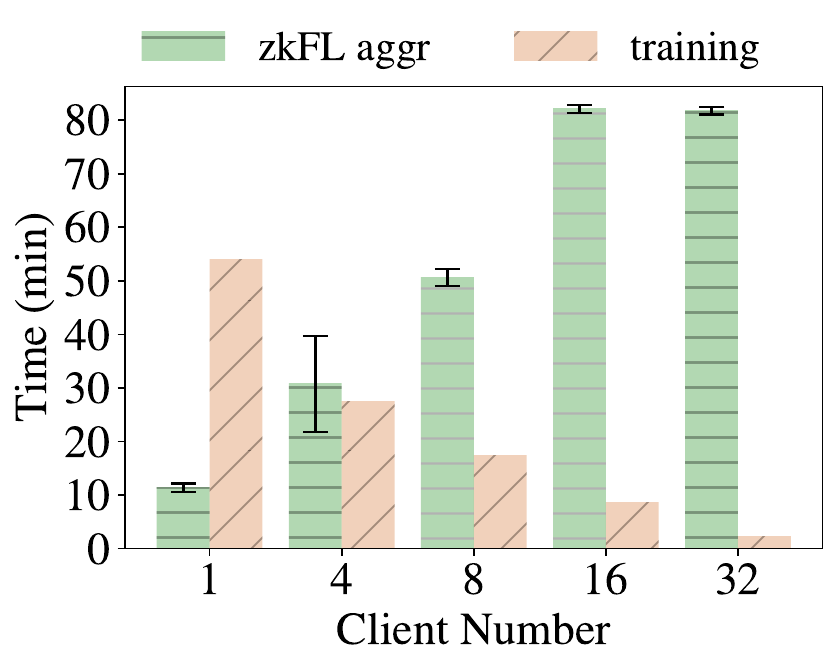}
\label{fig: densenet201_aggregation_time}
}%
\centering
\caption{Accuracy, training, encryption, and aggregation time for Densenet201 on CIFAR-10 with \zkFL system.}
\label{fig:encryption}
\end{figure*}

\begin{figure*}[t]
\centering
\subfigure[Training perplexity.]{
\includegraphics[width=0.48\columnwidth]{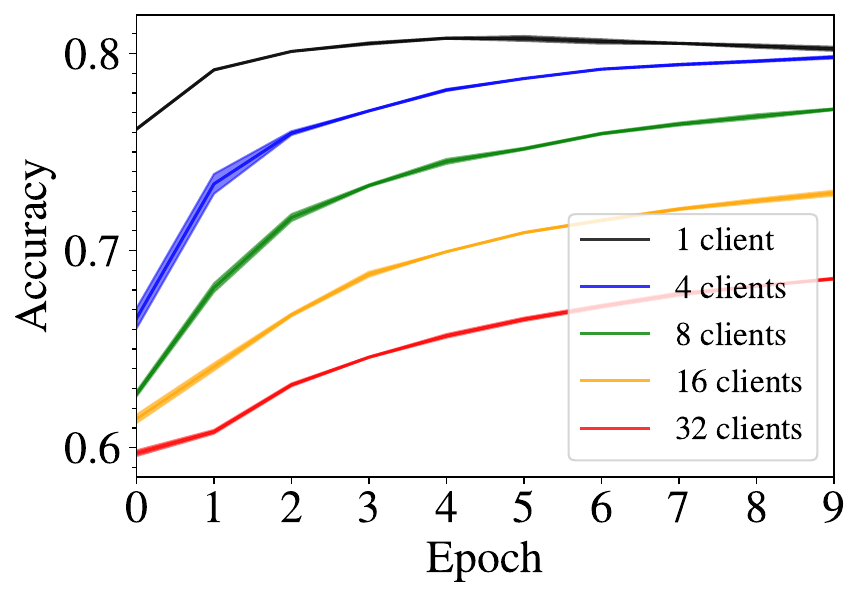}
\label{fig: 1L_accuracy_time}
}%
\subfigure[Total training time.]{
\includegraphics[width=0.48\columnwidth]{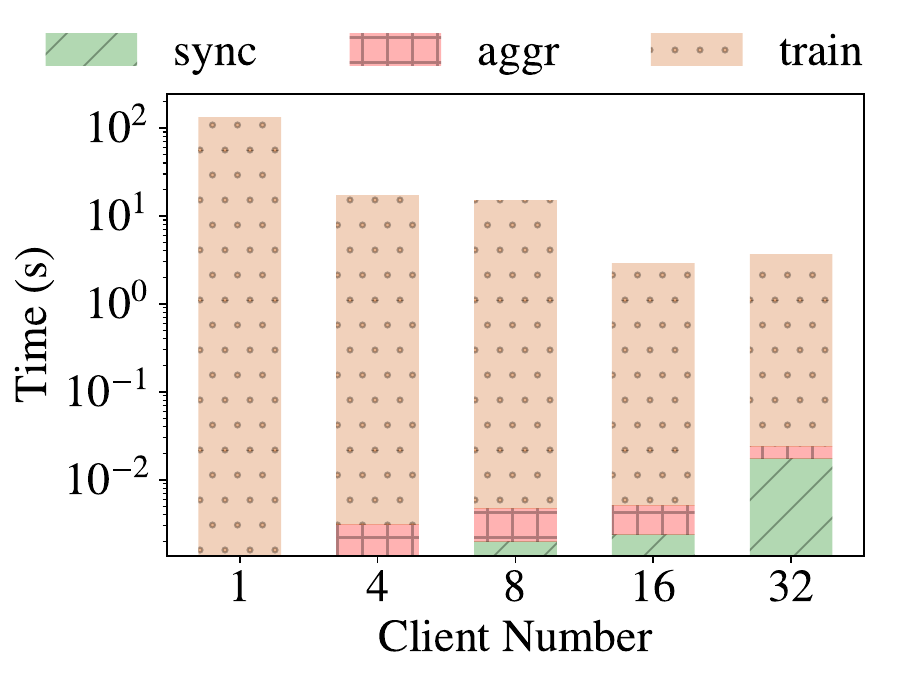}
\label{fig: 1L_train_time}
}%
\hfill
\subfigure[\zkFL encryption time.]{
\includegraphics[width=0.48\columnwidth]{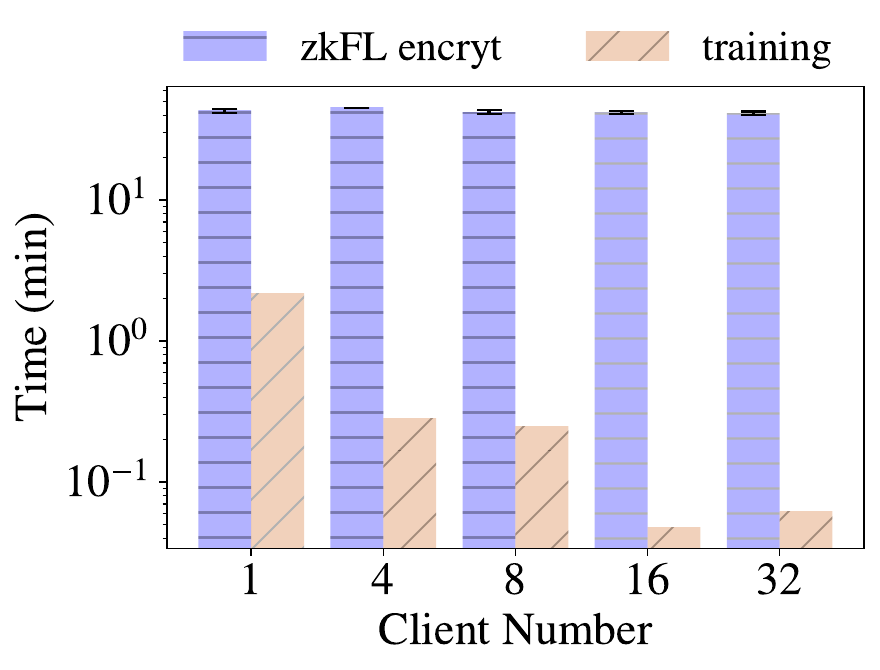}
\label{fig: 1L_encryption_time}
}%
\subfigure[\zkFL aggregation time.]{
\includegraphics[width=0.48\columnwidth]{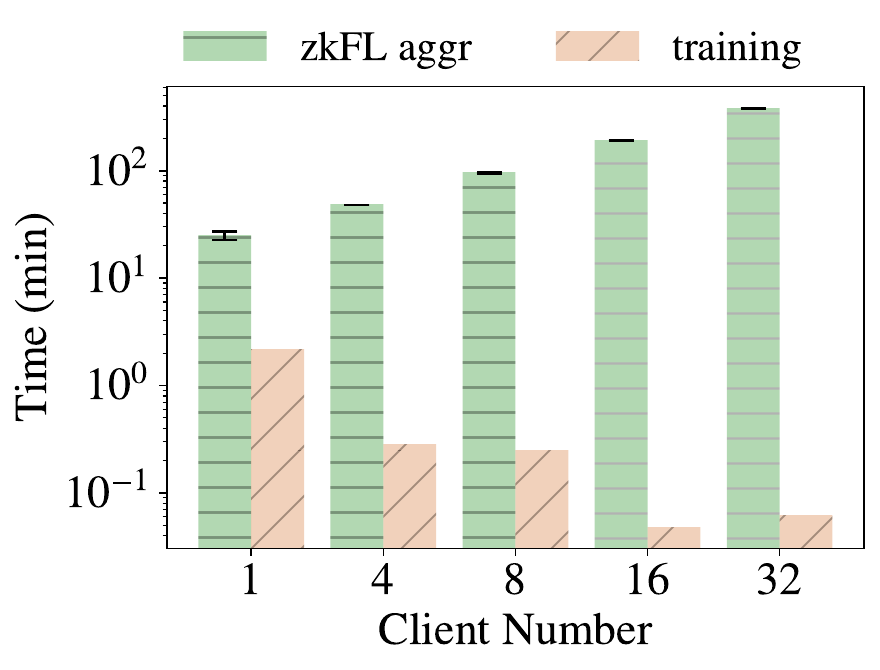}
\label{fig: 1L_aggregation_time}
}%
\centering
\caption{Perplexity, training, encryption, and aggregation time for LSTM (one layer) on PTB with \zkFL system.}
\end{figure*}

\begin{figure*}[t]
\centering
\subfigure[Training perplexity.]{
\includegraphics[width=0.48\columnwidth]{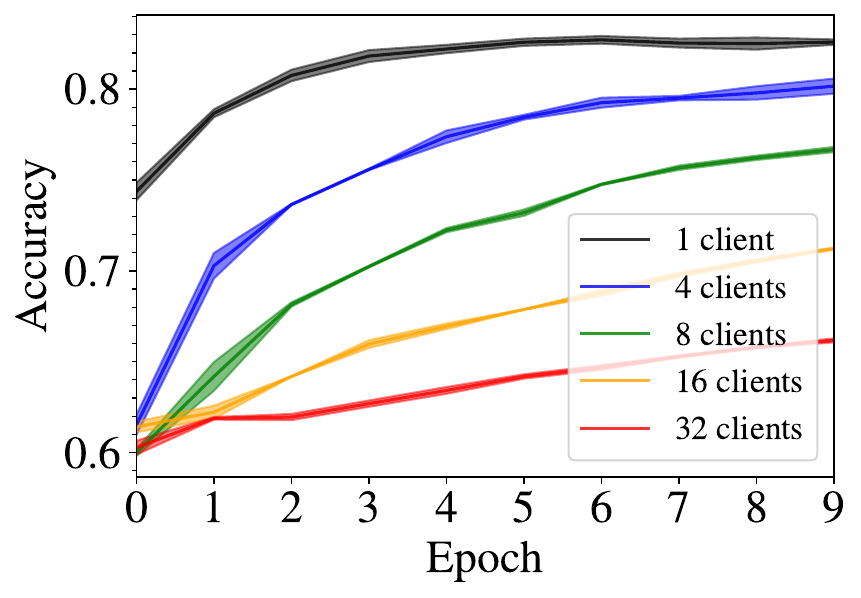}
\label{fig: 2L_accuracy_time}
}%
\subfigure[Total training time.]{
\includegraphics[width=0.48\columnwidth]{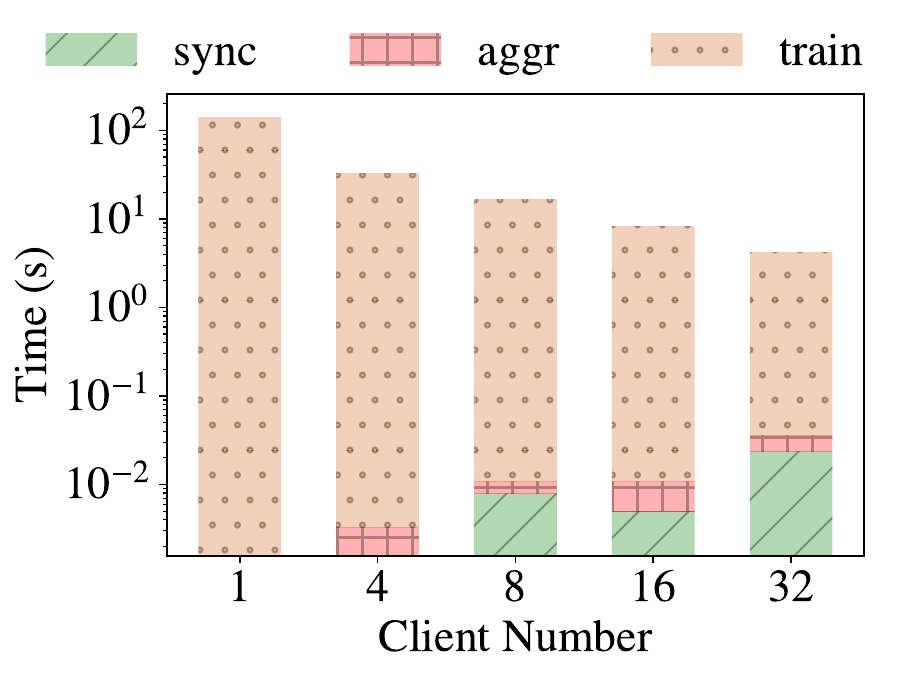}
\label{fig: 2L_train_time}
}%
\hfill
\subfigure[\zkFL encryption time.]{
\includegraphics[width=0.48\columnwidth]{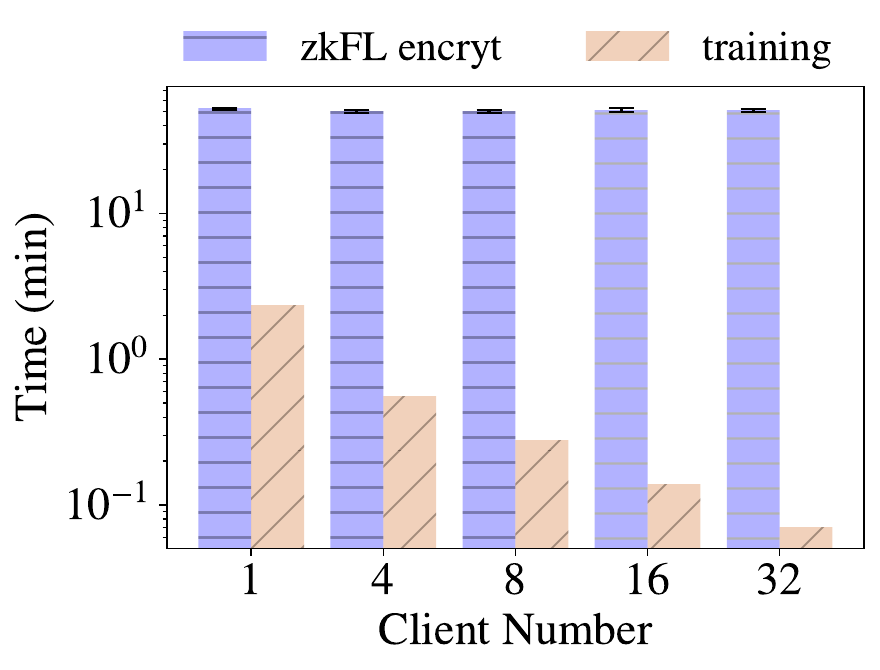}
\label{fig: 2L_encryption_time}
}%
\subfigure[\zkFL aggregation time.]{
\includegraphics[width=0.48\columnwidth]{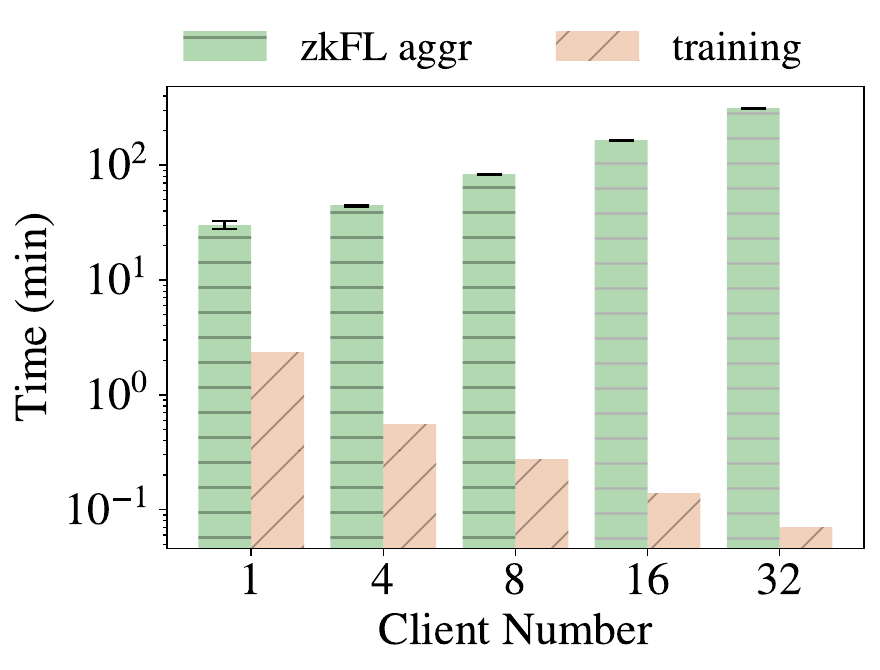}
\label{fig: 2L_aggregation_time}
}%
\centering
\caption{Perplexity, training, encryption, and aggregation time for LSTM (two layers) on PTB with \zkFL system.}
\end{figure*}

\begin{figure*}[t]
\centering
\subfigure[Training perplexity.]{
\includegraphics[width=0.48\columnwidth]{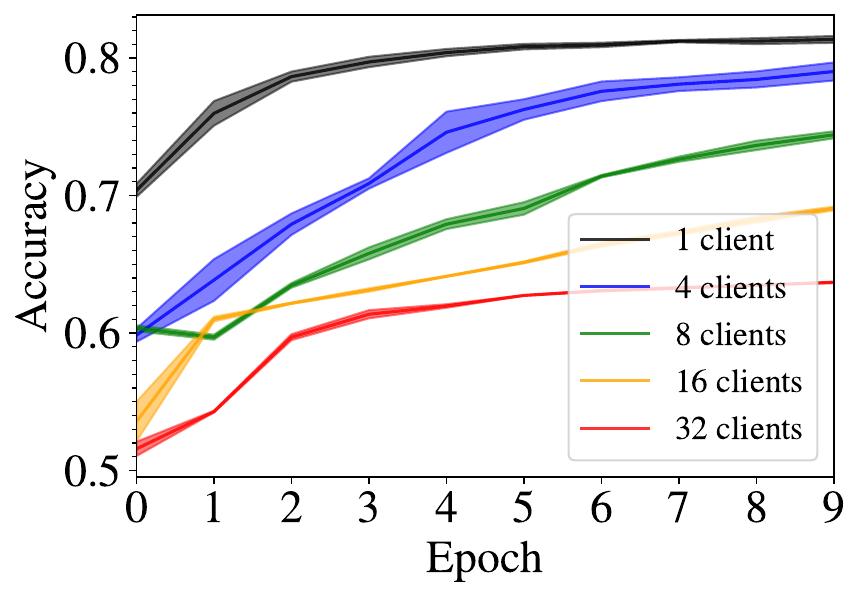}
\label{fig: 3L_accuracy_time}
}%
\subfigure[Total training time.]{
\includegraphics[width=0.48\columnwidth]{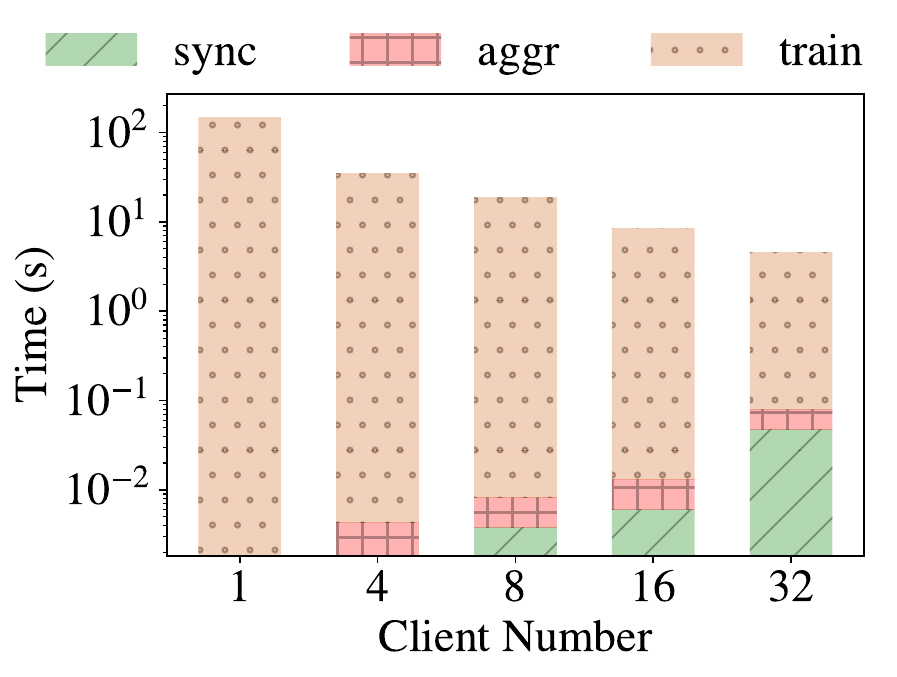}
\label{fig: 3L_train_time}
}%
\hfill
\subfigure[\zkFL encryption time.]{
\includegraphics[width=0.48\columnwidth]{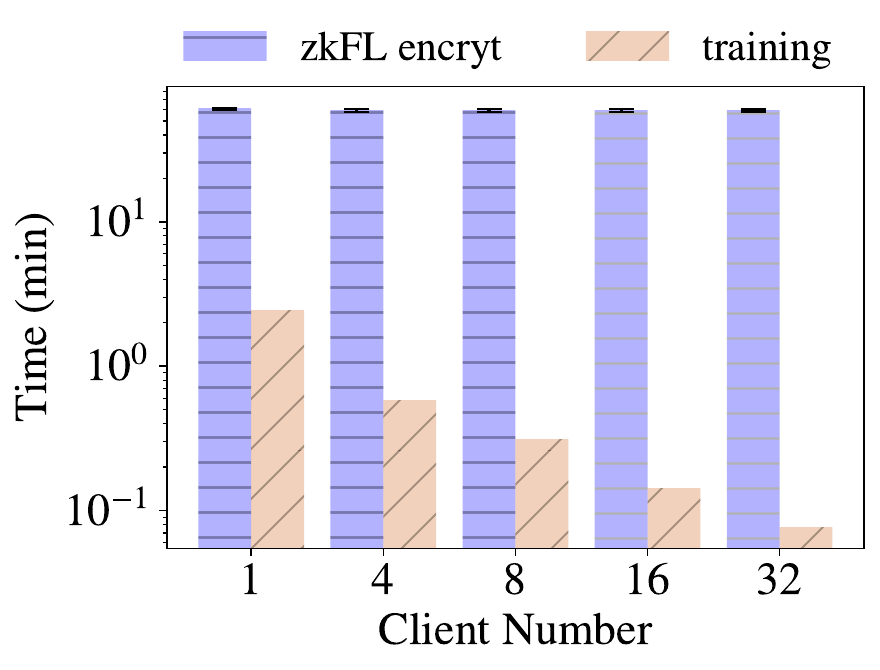}
\label{fig: 3L_encryption_time}
}%
\subfigure[\zkFL aggregation time.]{
\includegraphics[width=0.48\columnwidth]{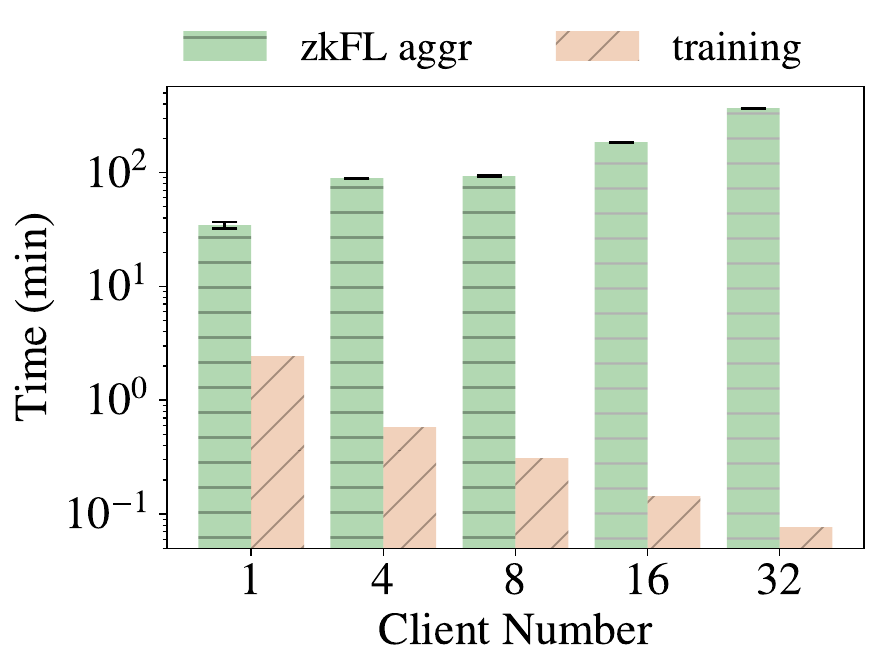}
\label{fig: 3L_aggregation_time}
}%
\centering
\caption{Perplexity, training, encryption, and aggregation time for LSTM (three layers) on PTB with \zkFL system.}
\end{figure*}

\subsubsection{Data and Task} We consider two benchmark machine learning tasks under the common federated setup. We use FedAVG~\cite{mcmahan2017communication} as the base \FL method to evaluate \zkFL. In each epoch, clients train their models separately with local data. Then, the aggregator synchronizes local models and performs the model evaluation. The training set is split into $K$ subsets of equal size and distributed across $K$ clients. For each model of interest, we conduct the tasks for \zkFL with $1$, $4$, $8$, $16$, and $32$ client(s). For each task, we record the training time without using \zkFL, the encryption and aggregation time with \zkFL, as well as the \ZKP generation and verification time under \zkFL. We also evaluate the performance of each task with various network backbones and client numbers.

\begin{itemize}
    \item \textbf{Image Classification}: We consider image classification on CIFAR-10 dataset, a benchmark computer vision task~\cite{dai2019toward}. We use the default train-test split of CIFAR-10. For each client, $10\%$ of distributed data is assigned as the validation set. We use a standard Adam optimizer~\cite{kingma2015adam} with fixed learning rate 0.001 and batch size 50. We test our \zkFL system with two families of network architectures: ResNets~\cite{he2016deep} (ResNet18, ResNet34, and ResNet50) and DenseNets~\cite{huang2017densely} (DenseNet121, DenseNet169, and DenseNet201). We use this setup to evaluate the sensitivity of \zkFL to network architectures and number of parameters. We use the \emph{area under the receiver operating curve} (AUROC) to evaluate the accuracy.
    \item \textbf{Language Understanding}: We consider language modeling (word prediction) on the Penn Treebank (PTB) dataset, a benchmark natural language processing task~\cite{dai2019toward}. We use the default train-validation-test split. We test our zkFL system with three different network architectures. We use long short-term memory (LSTM) network~\cite{hochreiter1997long} to model the language. Each LSTM consists of an embedding layer, a few LSTM layers, and a linear layer. To perform a similar sensitivity analysis of \zkFL as above, we consider LSTMs with one, two, and three LSTM layers, with 650 neurons in each layer. We use a standard Adam optimizer with fixed learning rate 0.001 and batch size 20. The dropout ratio is 0.5.
\end{itemize}

\subsubsection{Implementation} We develop a prototype of \zkFL. We implement the prototype in \href{https://github.com/arkworks-rs}{\texttt{Rust}} and interface it with \texttt{Python} to train deep learning models with PyTorch 1.13. We adopt the elliptic curve Curve25519 (\emph{\emph{i.e.}} 126-bit security) implementation from the \texttt{dalek} \href{https://github.com/dalek-cryptography/curve25519-dalek}{curve25519 library} for cryptographic encryption operations. We build Pedersen commitments~\cite{aranha2022eclipse} over the elliptic curves and integrate it with Halo2~\cite{bowe2019recursive}, a \ZKP system that is being used by the \href{https://github.com/zcash/halo2/}{\texttt{Zcash}}. All tests are implemented in PyTorch 1.10.1 on an NVIDIA Tesla T4 GPU with 16GB memory.

\subsection{Results of \zkFL}

\subsubsection{Training Time} We commence our evaluation by conducting an in-depth analysis of the total training time per epoch for each network backbone, focusing on the conventional FL approach without the integration of \zkFL. This evaluation encompasses three crucial components that contribute to the overall training time: the local training time of each individual client, the time required for aggregating local model updates on the central aggregator, and the {synchronization time}. As shown in Figs.~\ref{fig: resnet18_train_time}~\ref{fig: resnet34_train_time}~\ref{fig: resnet50_train_time}~\ref{fig: densenet121_train_time}~\ref{fig: densenet169_train_time}~\ref{fig: densenet201_train_time}~\ref{fig: 1L_train_time}~\ref{fig: 2L_train_time}~\ref{fig: 3L_train_time}, our findings indicate that the local training time of each client is the primary contributor to the total training time. Moreover, as the number of clients increases, the local training time for individual clients decreases due to the effective distribution of training tasks among them.

\subsubsection{Encryption and Aggregation Time} We conduct a thorough evaluation of the encryption time for clients and the aggregation time for the central aggregator within the \zkFL system. In addition to the computational costs associated with the FL training protocol, the client-side tasks involve computing commitments, such as encryption computations for each model update parameter. Figs~\ref{fig: resnet18_encryption_time}~\ref{fig: resnet34_encryption_time}~\ref{fig: resnet50_encryption_time}~\ref{fig: densenet121_encryption_time}~\ref{fig: densenet169_encryption_time}~\ref{fig: densenet201_encryption_time}~\ref{fig: 1L_encryption_time}~\ref{fig: 2L_encryption_time}~\ref{fig: 3L_encryption_time}
demonstrate that this additional cost varies based on the choice of underlying network backbones and increases with the number of parameters in the network. For instance, the encryption time for ResNet18 (\emph{\emph{i.e.}}, $10 - 20$ mins) is lower than that for ResNet34 (\emph{\emph{i.e.}}, $25 - 45$ mins), as the latter has a larger number of network parameters. Moreover, as illustrated in Figs~\ref{fig: resnet18_aggregation_time}~\ref{fig: resnet34_aggregation_time}~\ref{fig: resnet50_aggregation_time}~\ref{fig: densenet121_aggregation_time}~\ref{fig: densenet169_aggregation_time}~\ref{fig: densenet201_aggregation_time}~\ref{fig: 1L_aggregation_time}~\ref{fig: 2L_aggregation_time}~\ref{fig: 3L_aggregation_time}, the aggregation time of the central aggregator in \zkFL is influenced by the network parameters and exhibits an approximately linear relationship with the number of clients in the \FL system, which has a critical effect on the whole system's efficiency.

\subsubsection{\ZKP Proof Generation and Verification Time} 
\begin{figure}[t]
\centering
\includegraphics[width=0.9\columnwidth]{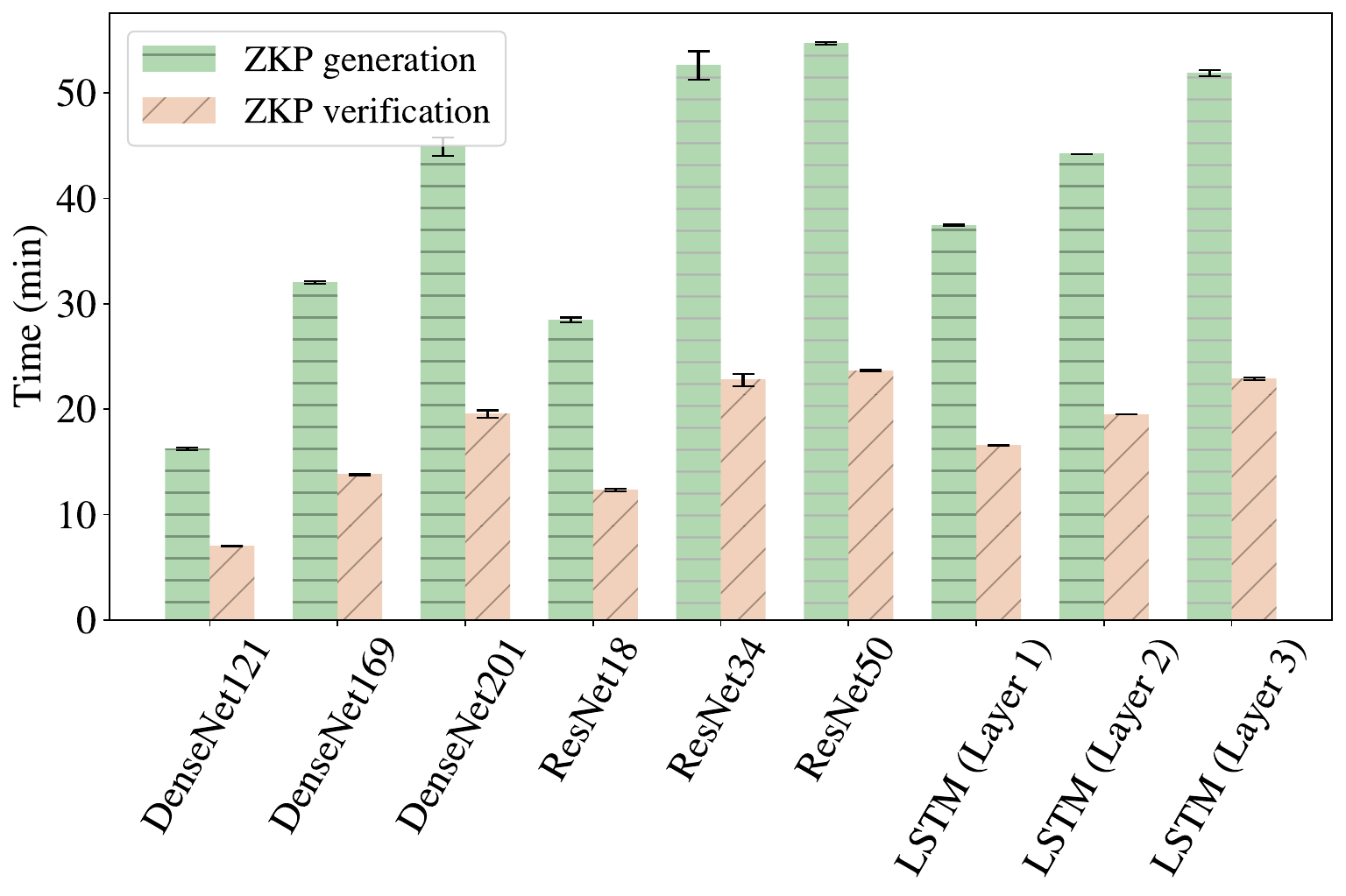}
\caption{Halo2 ZKP proof generation and verification time for a \zkFL system with various network backbones.}
\label{fig:halo2_proof_time}
\end{figure}

As depicted in Fig.~\ref{fig:halo2_proof_time}, the time required for Halo2 \ZKP generation and verification exhibits variations depending on the chosen network and increases with the size of the network parameters. Among the six networks evaluated, ResNet50 demands the longest time for proof generation, taking approximately $54.72\pm 0.12$ mins. Notably, the proof verification time is approximately half of the generation time. This favorable characteristic makes \zkFL more practical, as the aggregator, equipped with abundant computing power, can efficiently generate the proof, while the source-limited clients can verify it without significant computational overhead. This highlights the feasibility and applicability of \zkFL in real-world scenarios, where the \FL clients may have constrained resources compared to the central aggregator.

\subsubsection{Communication Costs} Compared to the traditional \FL, \zkFL will also increase the communication costs for the clients and the aggregator, as the encrypted local training model updates $Enc(w_i)\thickspace(1\leq i \leq n)$ and the \ZKP proof $\pi$. As shown in Table~\ref{tab: communication-costs}, for each network backbone, we compare the size of encrypted data with the size of the model updates in plaintext. Compared to traditional \FL, \zkFL will cause additional communication costs (e.g., the encrypted data). We estimate the proof size based on the network parameter size and the data provided in the Halo2 original paper~\cite{bowe2019recursive}.  We show that the size of encrypted data grows linearly in the number of parameters. Thus, the communication costs are dominated by the encrypted data. For ResNet50, the network backbone with the largest number of model parameters in our experiment, the additional data transmitted to the centralized aggregator per client is approximately {$2 \times$} compared to the plaintext). However, it is of utmost importance to underscore that although the relative size increase might appear significant, the absolute size of the updates is merely equivalent to a few minutes of compressed HD video. As a result, the data size remains well within the capabilities of the \FL clients utilizing wireless or mobile connections. For instance, let us examine the most substantial communication overhead as outlined in Table~\ref{tab: communication-costs}, specifically, a total data transfer volume of 491MB + $2 \times$491MB + 628KB = 1,473.61MB, associated with the network architecture of ResNet50. In a scenario where a client transmits model updates in plaintext, alongside the encrypted model updates and the \ZKP proof to a centralized aggregator, all over a network bandwidth of 1 Gbps, the resulting communication latency can be approximately calculated as follows:  $1{,}473.61 \times 8 /1000 = 11.79$ seconds. This latency is approximately $3\times$ of the one in \FL under the same communication network conditions.

\begin{table}[t]
\centering
\caption{{Estimated communication costs for \zkFL with various network backbones.} The increased communication costs in \zkFL are dominated by the encrypted (i.e., committed) data. }
\renewcommand\arraystretch{1.5}
    \resizebox{\columnwidth}{!}{ 
    \begin{tabular}{l|c|c|c|c}
        \toprule
        Models & \# Parameters   & \makecell{Model \\ Updates in \\ Plaintext} & \makecell{Estimated \\ Encrypted Model\\ Updates ($2\times$)} & \makecell{Estimated \\\ZKP Proof\\ Size} \\
        \midrule
        DenseNet121 &$\phantom{0}6{,}964{,}106$ &146MB &146MB&186KB \\
        DenseNet169 &$12{,}501{,}130$&262MB &262MB&334KB \\
        DenseNet201 &$18{,}112{,}138$&380MB &380MB&484KB\\
        \hline
        ResNet18 &$11{,}181{,}642$&238MB &238MB&299KB \\
        ResNet34 &$21{,}289{,}802$&452MB &452MB&569KB \\
        ResNet50 &$23{,}528{,}522$&497MB &497MB&628KB \\

        \hline
        LSTM (one layer) &$16{,}395{,}200$&374MB &374MB&438KB \\
        LSTM (two layers) &$19{,}780{,}400$&415MB &415MB&528KB \\
        LSTM (three layers) &$23{,}165{,}600$&486MB &486MB&619KB \\
        
        \bottomrule
    \end{tabular}
    }
    \label{tab: communication-costs}
\end{table}

\begin{figure*}[t]
\centering
\subfigure[Resnet18.]{
\includegraphics[width=0.312\linewidth]{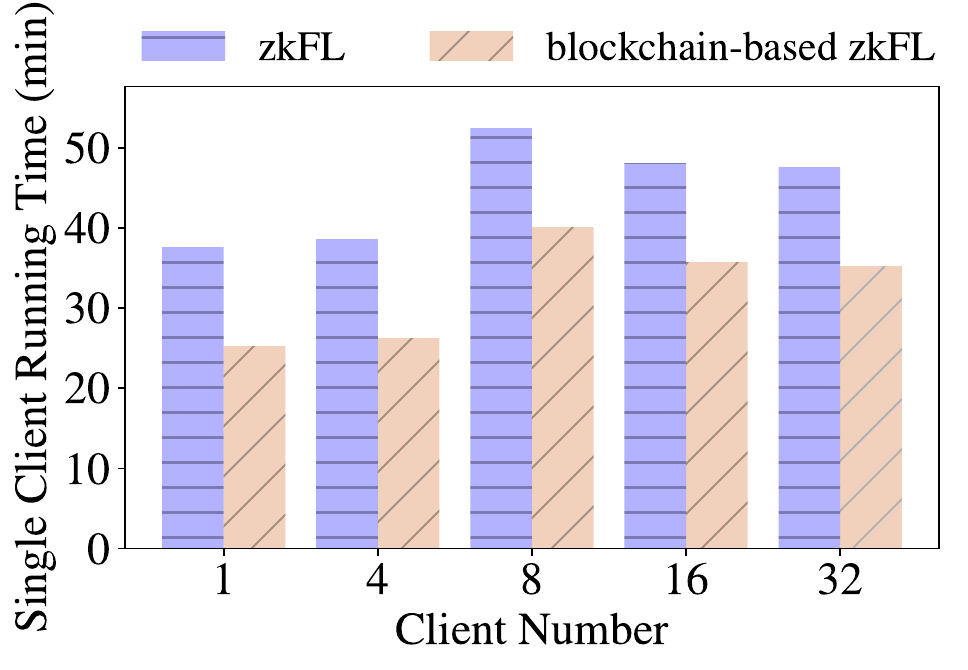}
\label{fig: resnet18_encrypt_traning_verify_time}
}%
\subfigure[Resnet34.]{
\includegraphics[width=0.312\linewidth]{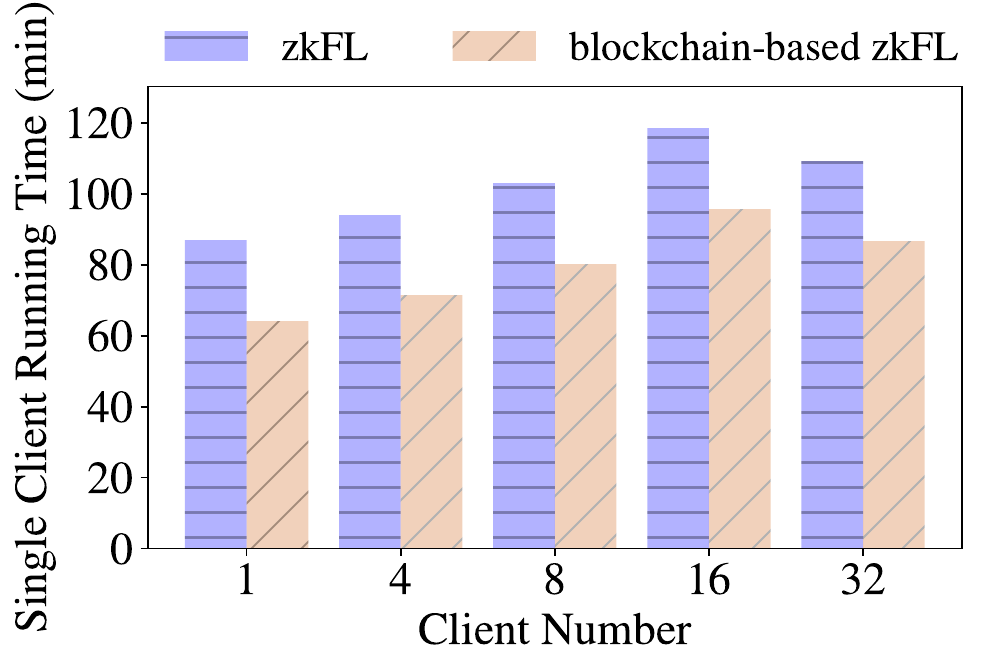}
\label{fig: resnet34_encrypt_traning_verify_time}
}%
\hfill
\subfigure[Resnet50.]{
\includegraphics[width=0.312\linewidth]{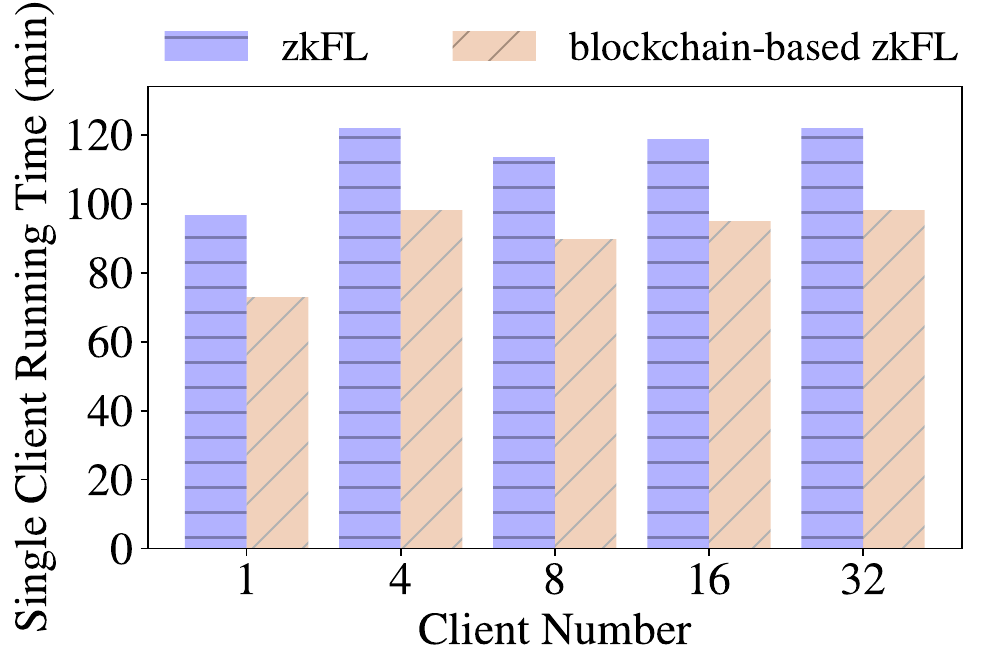}
\label{fig: resnet50_encrypt_traning_verify_time}
}%
\centering
\caption{Single client running time for Resnet18, Resnet34, and Resnet50 with \zkFL and blockchain-based \zkFL systems.}
\label{fig:Single-client-running-time-Resnet}
\end{figure*}

\begin{figure*}[t]
\centering
\subfigure[Densenet121.]{
\includegraphics[width=0.312\linewidth]{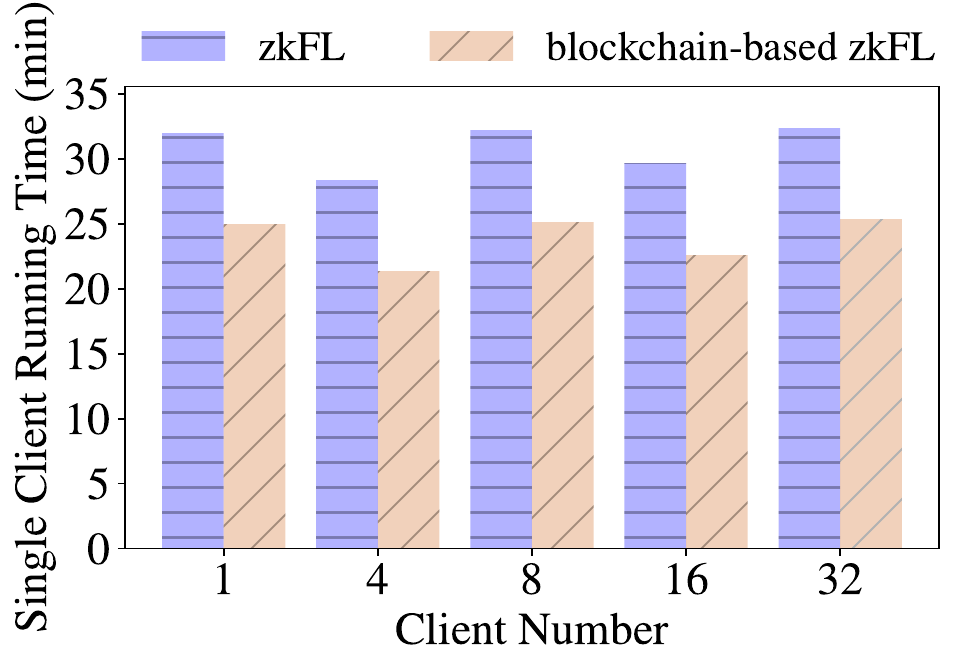}
\label{fig: densenet121_encrypt_traning_verify_time}
}%
\subfigure[Densenet169.]{
\includegraphics[width=0.312\linewidth]{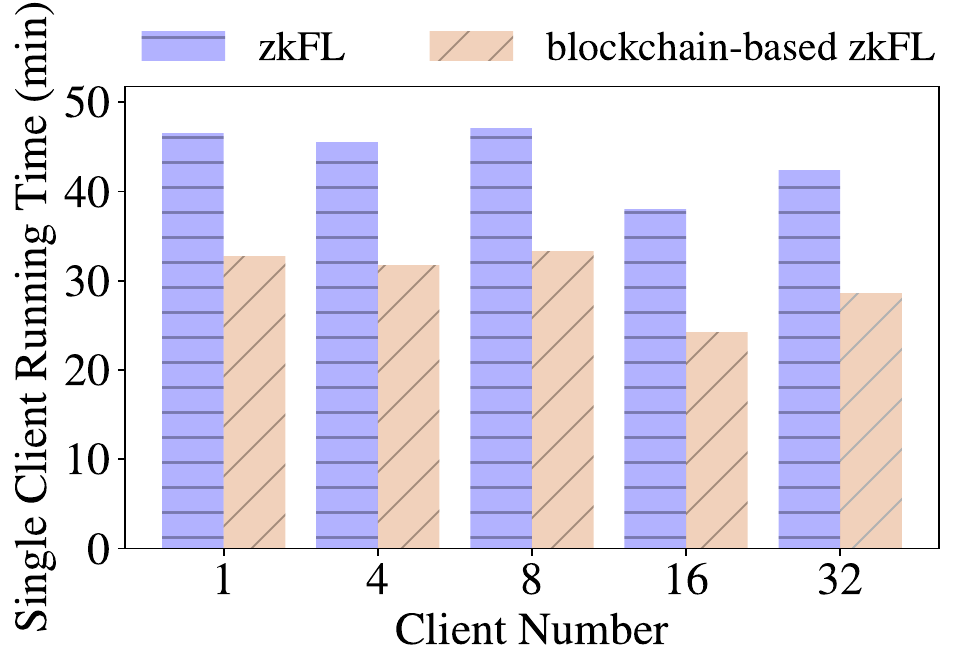}
\label{fig: densenet169_encrypt_traning_verify_time}
}%
\hfill
\subfigure[Densenet201.]{
\includegraphics[width=0.312\linewidth]{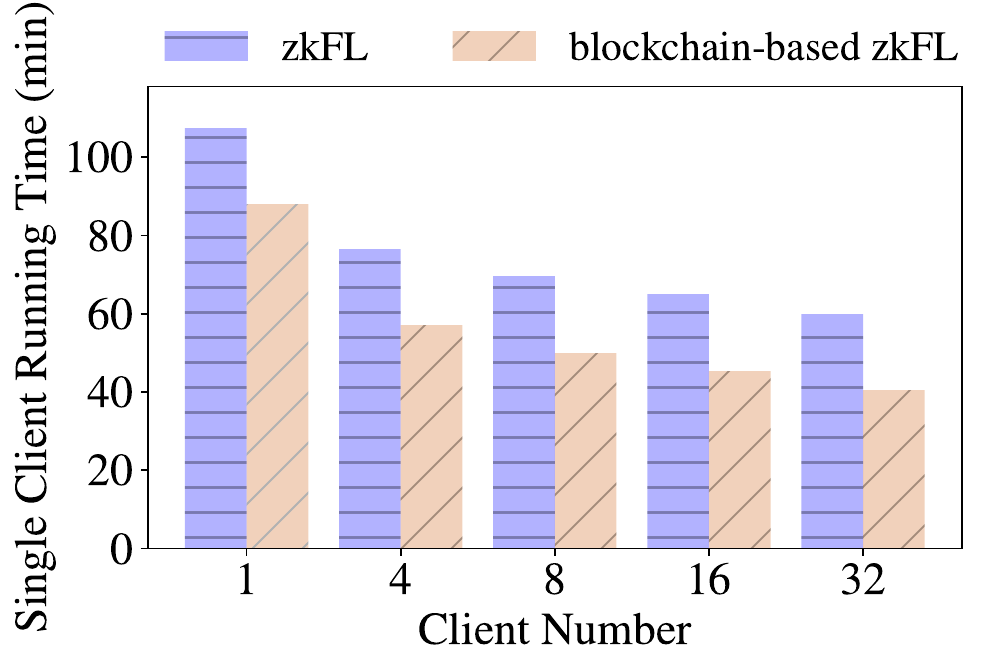}
\label{fig: densenet201_encrypt_traning_verify_time}
}%
\centering
\caption{Single client running time for Densenet121, Densenet169, and Densenet201 with \zkFL and blockchain-based \zkFL systems.}
\label{fig:Single-client-running-time-Densenet}
\end{figure*}

\begin{figure*}[htbp]
\centering
\subfigure[LSTM Layer 1.]{
\includegraphics[width=0.312\linewidth]{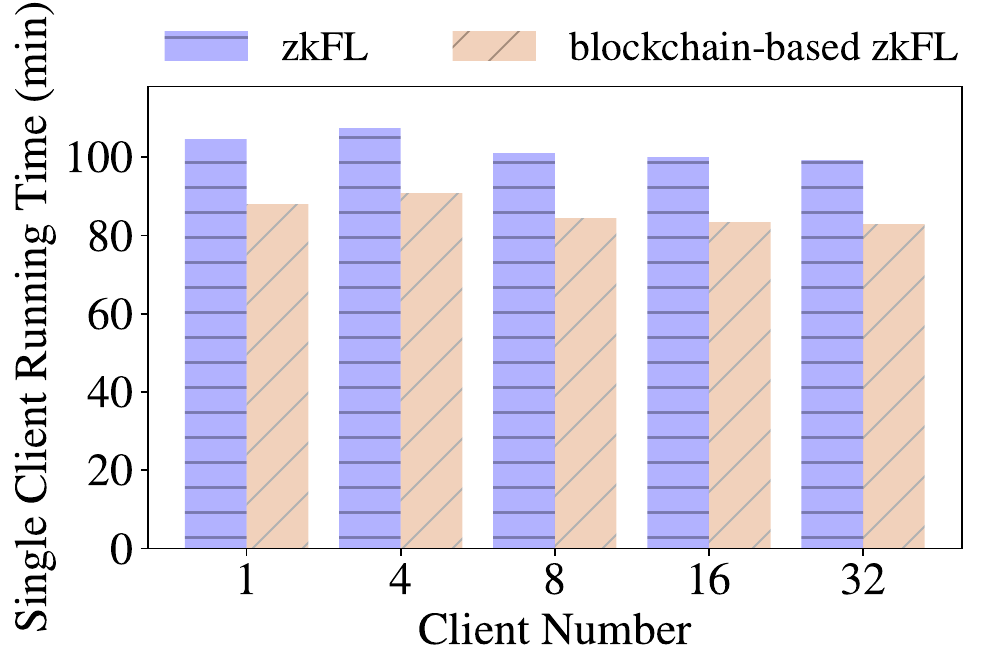}
\label{fig: 1L_encrypt_traning_verify_time}
}%
\subfigure[LSTM Layer 2.]{
\includegraphics[width=0.312\linewidth]{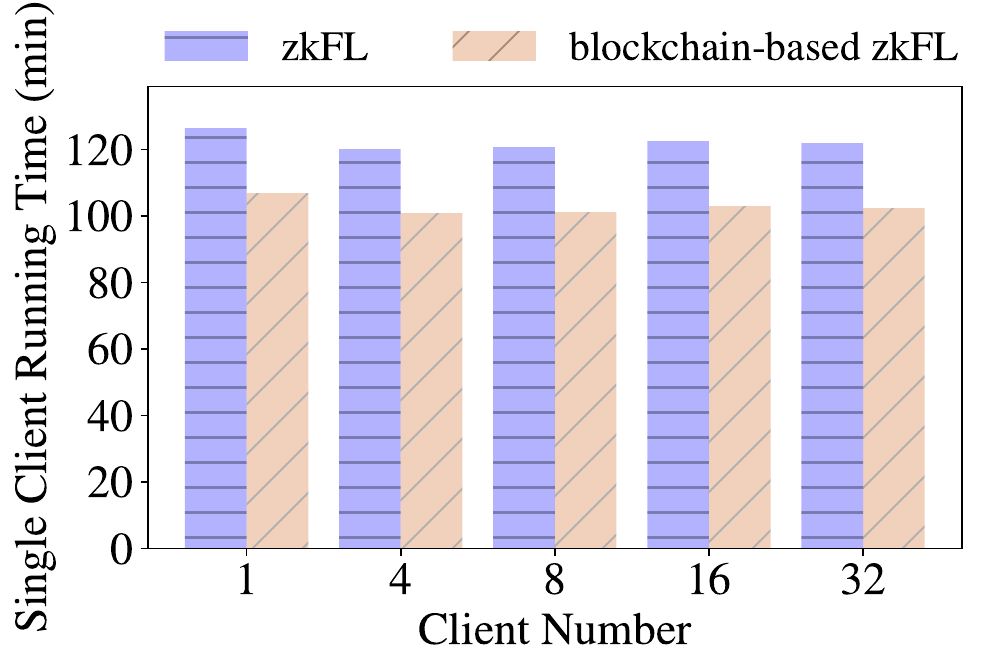}
\label{fig: 2L_encrypt_traning_verify_time}
}%
\hfill
\subfigure[LSTM Layer 3.]{
\includegraphics[width=0.312\linewidth]{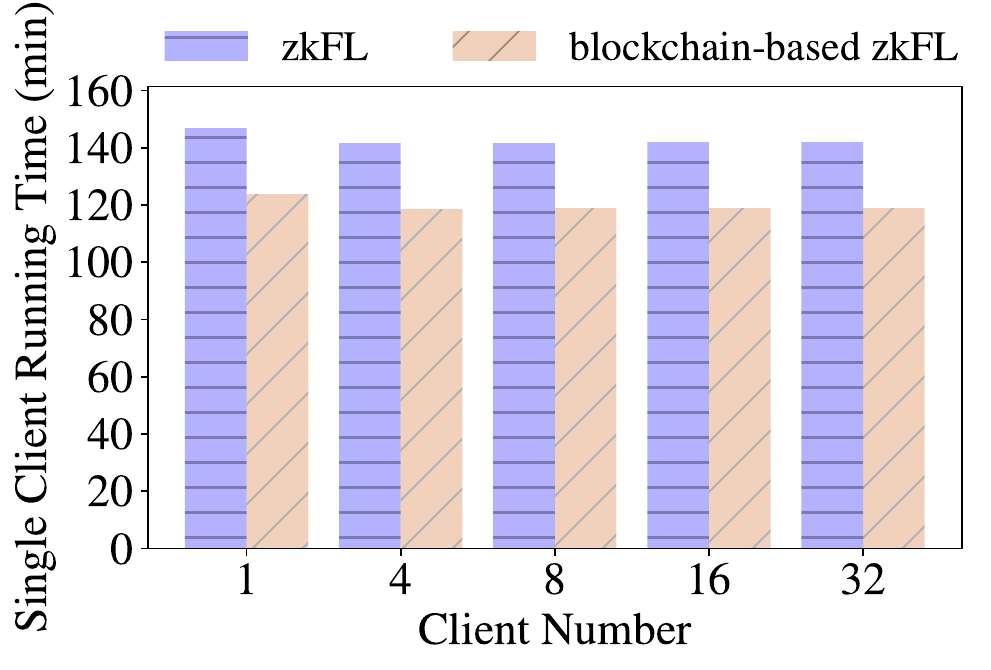}
\label{fig: 3L_encrypt_traning_verify_time}
}%
\centering
\caption{Single client running time for LSTM Layer 1, 2, and 3 with \zkFL and blockchain-based \zkFL systems.}
\label{fig:Single-client-running-time-LSTM}
\end{figure*}

\subsubsection{Training Performance and Convergence Analysis} In addition to analyzing the additional computation and communication costs introduced by \zkFL, we thoroughly investigate its potential impact on training performance, including accuracy and convergence speed. Theoretically, we demonstrate that compared to traditional \FL, \zkFL solely affects the data output from the clients, leaving the training process unaffected. Our experimental results provide strong evidence supporting this claim. Figs~\ref{fig: resnet18_accuracy_time}~\ref{fig: resnet34_accuracy_time}~\ref{fig: resnet50_accuracy_time}~\ref{fig: densenet121_accuracy_time}~\ref{fig: densenet169_accuracy_time}~\ref{fig: densenet201_accuracy_time}~\ref{fig: 1L_accuracy_time}~\ref{fig: 2L_accuracy_time}~\ref{fig: 3L_accuracy_time} present the final training results' accuracy/perplexity and convergence speed for both traditional \FL and \zkFL. We observe the accuracy and speed do not exhibit any difference between \FL settings with and without \zkFL in terms of epoch number. Furthermore, we observe that the convergence speed is primarily influenced by the number of clients involved in the process. This reaffirms the practical viability of \zkFL as it maintains training performance while enhancing security and privacy in the federated learning framework.

\subsection{Results of Blockchain-based \zkFL}

In this subsection, we present the results to show how blockchain-based \zkFL affects the performance of the system.

\subsubsection{Single Client Running Time}
To understand the efficiency gains of blockchain-based \zkFL over traditional \zkFL, we first focus on the average runtime for a single client. As depicted in Figs.~\ref{fig:Single-client-running-time-Resnet}, \ref{fig:Single-client-running-time-Densenet}, and~\ref{fig:Single-client-running-time-LSTM}, blockchain-based \zkFL shows reduced client running time. As detailed in Section~\ref{sec: Efficiency-Analysis}, this improvement stems from blockchain-based \zkFL clients not having to verify ZKP proofs produced by the aggregator. Instead, blockchain miners undertake the verification, allowing clients to simply access the validated data, specifically the hash of the encrypted aggregated model updates.

\begin{figure*}[t]
\centering
\subfigure[Resnet18.]{
\includegraphics[width=0.312\linewidth]{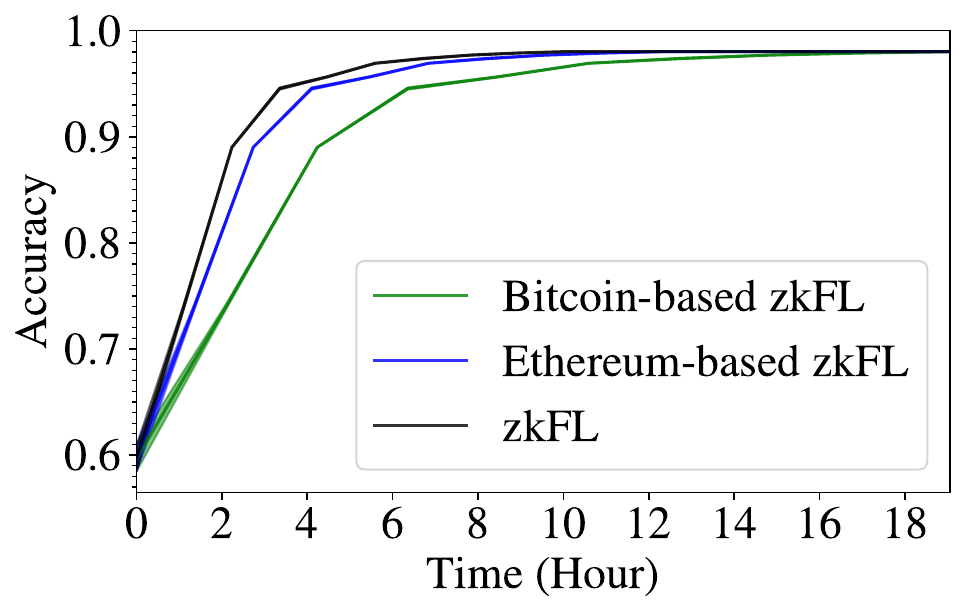}
\label{fig: resnet18_train_accuracy}
}%
\subfigure[Resnet34.]{
\includegraphics[width=0.312\linewidth]{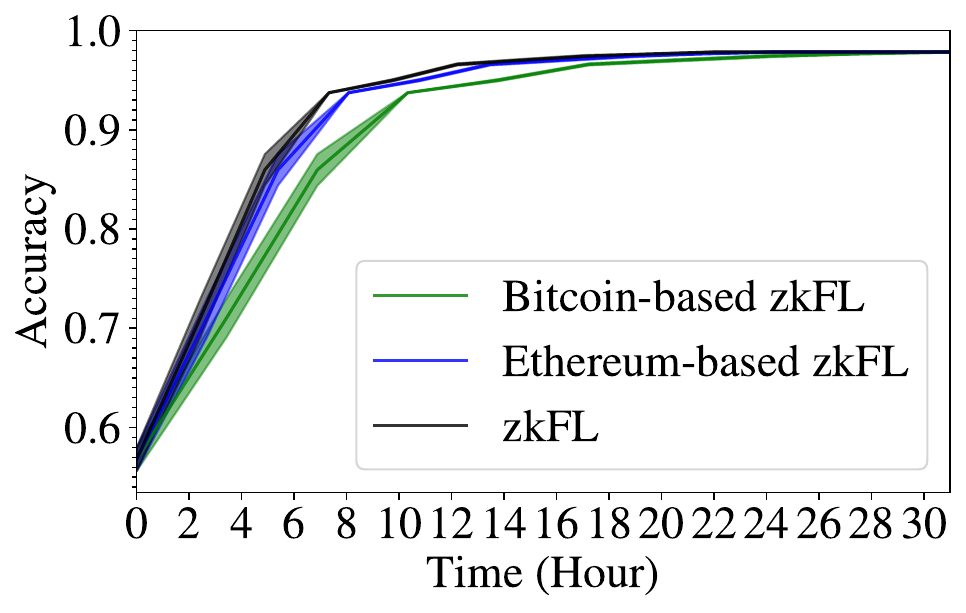}
\label{fig: resnet34_train_accuracy}
}%
\hfill
\subfigure[Resnet50.]{
\includegraphics[width=0.312\linewidth]{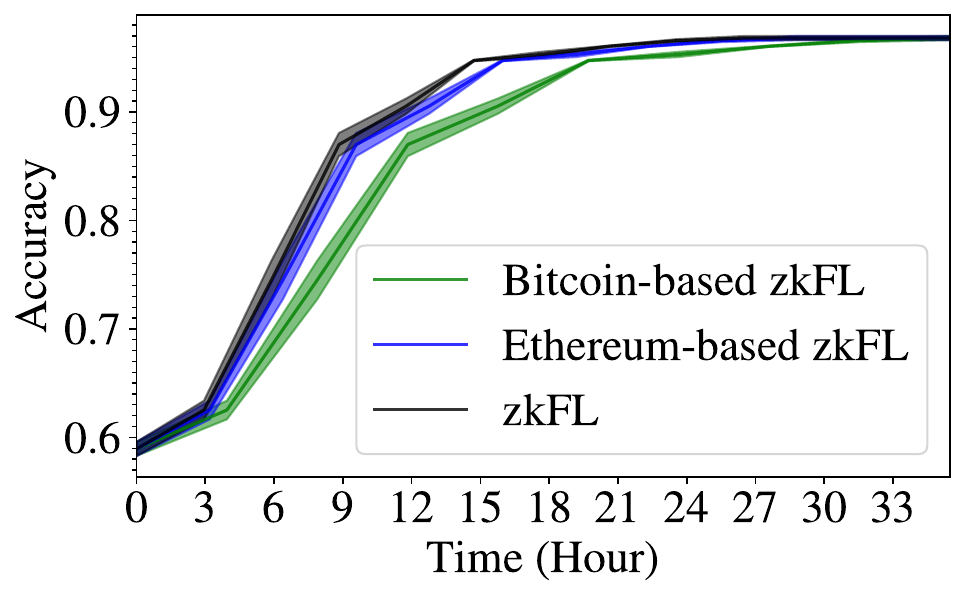}
\label{fig: resnet50_train_accuracy}
}%
\centering
\caption{Training accuracy over time for Resnet18, Resnet34, and Resnet50 with \zkFL and blockchain-based \zkFL systems with 4 clients.}
\label{fig:train-accuracy-time-Resnet}
\end{figure*}

\begin{figure*}[t]
\centering
\subfigure[Densenet121.]{
\includegraphics[width=0.312\linewidth]{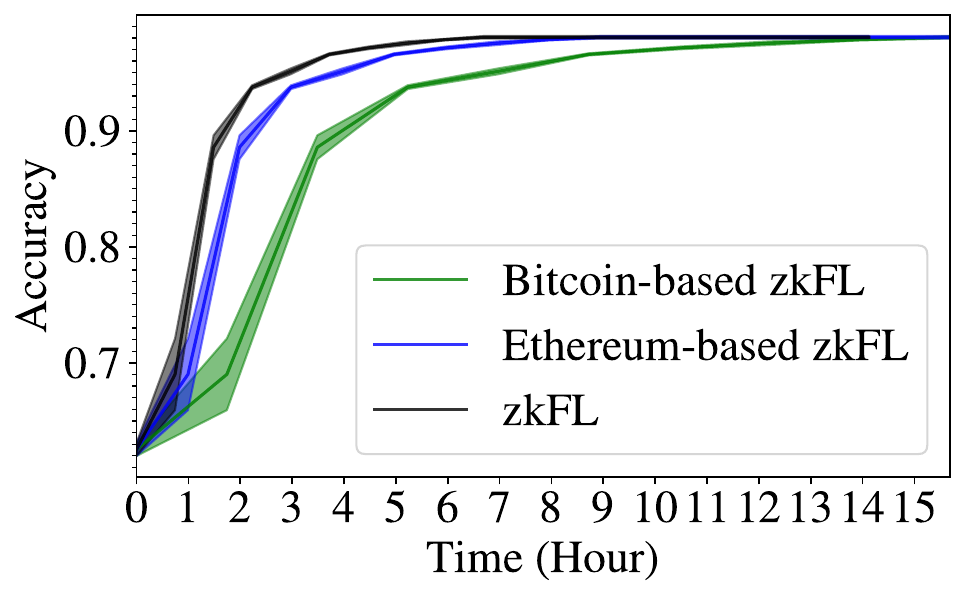}
\label{fig: densenet121_train_accuracy}
}%
\subfigure[Densenet169.]{
\includegraphics[width=0.312\linewidth]{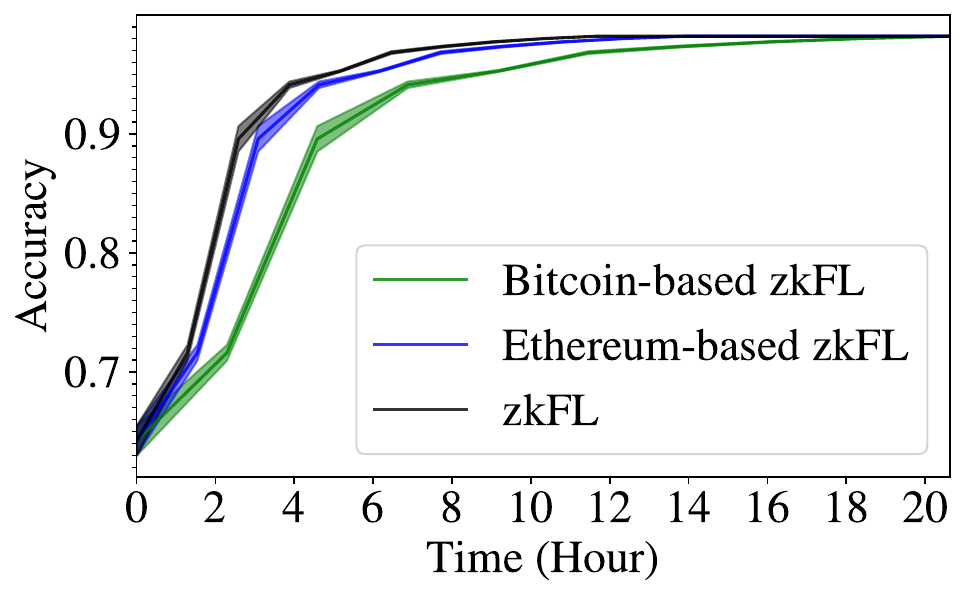}
\label{fig: densenet169_train_accuracy}
}%
\hfill
\subfigure[Densenet201.]{
\includegraphics[width=0.312\linewidth]{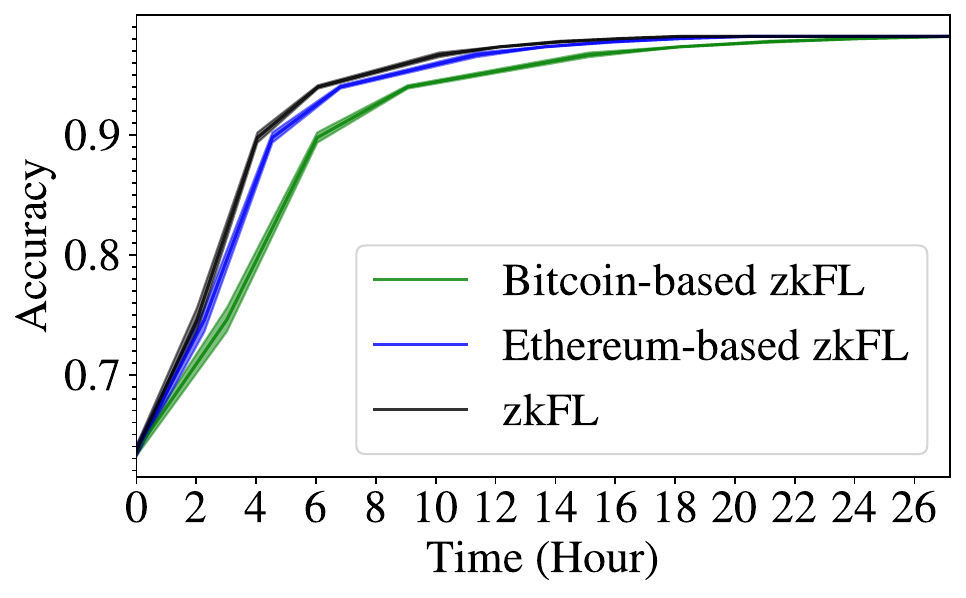}
\label{fig: densenet201_train_accuracy}
}%
\centering
\caption{Training accuracy over time for Densenet121, Densenet169, and Densenet201 with \zkFL and blockchain-based \zkFL systems with 4 clients.}
\label{fig:train-accuracy-time-Densenet}
\end{figure*}

\begin{figure*}[t]
\centering
\subfigure[LSTM (one layer).]{
\includegraphics[width=0.312\linewidth]{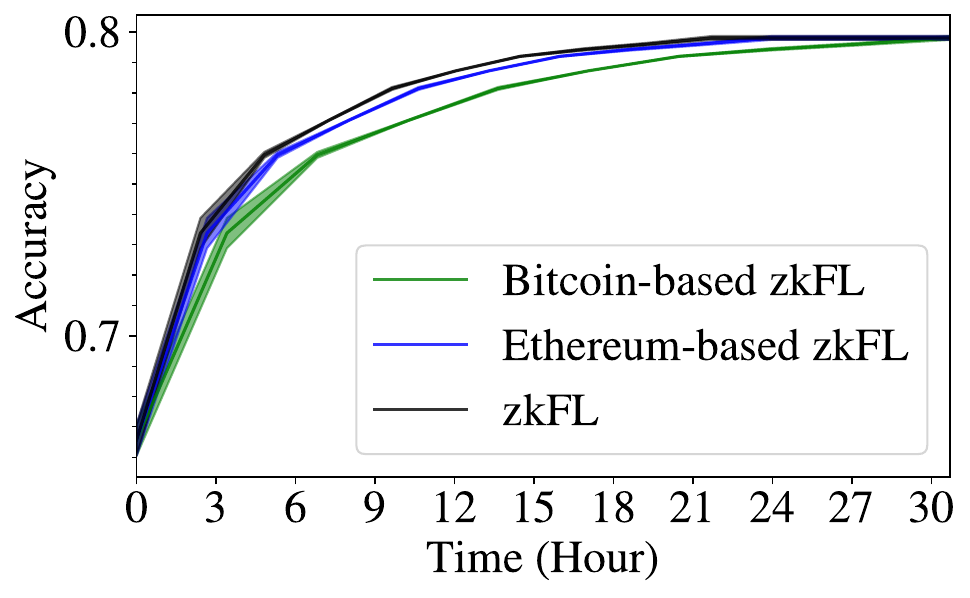}
\label{fig: 1L_train_accuracy}
}%
\subfigure[LSTM (two layers).]{
\includegraphics[width=0.312\linewidth]{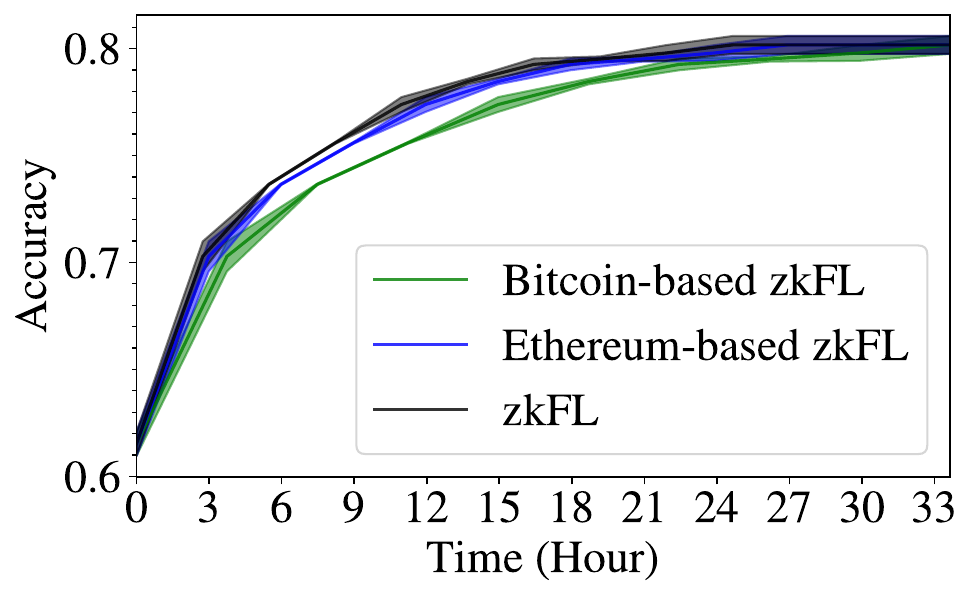}
\label{fig: 2L_train_accuracy}
}%
\hfill
\subfigure[LSTM (three layers).]{
\includegraphics[width=0.312\linewidth]{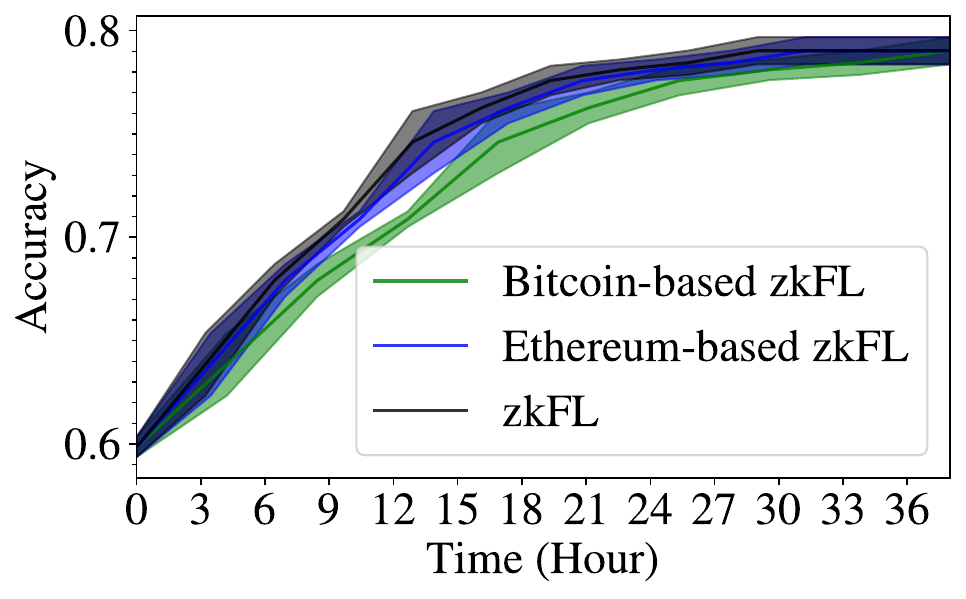}
\label{fig: 3L_train_accuracy}
}%
\centering
\caption{Training perplexity over time for LSTMs on PTB with \zkFL and blockchain-based \zkFL systems with 4 clients.}
\label{fig:train-accuracy-time-LSTM}
\end{figure*}

\subsubsection{Training Security, Performance and Convergence Analysis} We then analyze how blockchain-based \zkFL will affect the training convergence of \FL. The clients in a blockchain-based \zkFL rely on the blockchain miners to verify the \ZKP proofs and then append the hash value of the encrypted aggregated model update, $H(Enc(w))$, into the blockchain. To ensure that the miners have correctly performed the verification, the clients need to wait for the transaction that contains $H(Enc(w))$ to be finalized, to guarantee the security of the system. We adopt the two prominent blockchains, Bitcoin and Ethereum, as examples. For Bitcoin, it takes approximately six blocks to ensure the finality of a transaction, which is about one hour~\cite{carlsten2016instability}. On Ethereum, it takes about $15$ minutes for a block to finalize\footnote{\url{https://ethereum.org/fil/roadmap/single-slot-finality}}. Given these parameters, We graphically represent the training accuracy over time for standard \zkFL, alongside its Bitcoin and Ethereum counterparts. Figs~\ref{fig:train-accuracy-time-Resnet}, \ref{fig:train-accuracy-time-Densenet}, and~\ref{fig:train-accuracy-time-LSTM} demonstrate that, despite the delays caused by the blockchain transaction finalization, blockchain-based \zkFL achieves convergence in training accuracy, analogous to \zkFL without blockchain.

\subsubsection{On-Chain Costs}  
% In the design of \cite{dong2024defending}, the aggregation process is performed on-chain, and the aggregated model is also stored on-chain, resulting in an on-chain computation cost of at least $O(n\cdot m)$ and storage cost of $O(m)$, where $n$ is the number of clients and $m$ is the size of aggregated model. However, in our blockchain-based \zkFL design, the aggregator performs the aggregation off-chain and only stores its hash value on-chain, leading to a constant cost of $O(1)$. We consider the case of a model with a size of $1$MB and the hash function \texttt{SHA256}. According to the specifications of Ethereum~\cite{wood2014ethereum}, it requires $20$K gas\footnote{Gas is the pricing value required to conduct a transaction or execute a contract on Ethereum.} to save a 256-bit word to storage. Thus in \cite{dong2024defending}, it needs at least $1\times 10^6 \times 8 /256 \times 20 \times 10^3 = 625$M gas. In contrast, our blockchain-based \zkFL only requires $20$K gas. This difference becomes larger when we consider the cost for on-chain computation in \cite{dong2024defending}.
To compare the scalability of our blockchain-based \zkFL with other blockchain-based FL such as~\cite{dong2024defending}, we analyze their on-chain costs on smart-contract enabled blockchain, i.e., Ethereum. The design in~\cite{dong2024defending} involves performing the aggregation process on-chain and storing the aggregated model on-chain as well. This approach incurs an on-chain computation cost of at least \(O(n \cdot m)\) and a storage cost of \(O(m)\), where \(n\) represents the number of clients and \(m\) signifies the size of the aggregated model. Conversely, our blockchain-based \zkFL framework optimizes resource utilization by conducting the aggregation process off-chain and storing only the hash of the aggregated model on-chain, which reduces the cost to a constant \(O(1)\).

For a practical comparison, we consider a scenario where the model size is $1$MB, and the hash function employed is \texttt{SHA}-256. According to Ethereum's specifications \cite{wood2014ethereum}, storing a $256$-bit word requires $20{,}000$ gas. Therefore, the method described in~\cite{dong2024defending} would necessitate at least \(\frac{1 \times 10^6 \times 8}{256} \times 20{,}000 = 625\)M gas for storage alone. In stark contrast, our blockchain-based \zkFL requires a mere $20{,}000$ gas for storing the hash, highlighting a significant efficiency improvement. This disparity is further accentuated when accounting for the on-chain computation costs associated with \cite{dong2024defending}.

\section{Discussion}\label{sec: discussion}
In the following, we discuss the limit of our \zkFL designs and potential future work for improvement.

\noindent\textbf{Decentralized Storage.} In our blockchain-based \zkFL design, the miners will only store the hash value of the encrypted aggregated model update $H(Enc(w))$ on-chain, rather than $Enc(w)$. This approach addresses the impracticality and high cost of storing large data on most existing blockchains.  Moreover, the clients can directly receive $w$ and $Enc(w)$ from the centralized aggregator. However, the $Enc(w)$ will be propagated to the blockchain miners and cause communication costs. To reduce the costs, we can leverage decentralized storage platforms, such as IPFS\footnote{https://ipfs.tech/}, or blockchains for decentralized storage, such as Filecoin~\footnote{https://filecoin.io/}. These platforms enable storage of large-sized encrypted model updates, accessible to miners without the need to broadcast them repeatedly across the blockchain's P2P network with which miners connect.

\smallskip \noindent\textbf{Recursive Proofs for ZKPs.} We have demonstrated that \zkFL enhances the security of traditional \FL at the expense of additional computation for \ZKP proof generation and verification. To mitigate these computational costs, recursive zero-knowledge proofs~\cite{kothapalli2022nova} could be utilized.  By employing recursion in ZKPs, complex computations can be broken down into smaller, more manageable sub-proofs. This is particularly advantageous in scenarios involving multiple layers of computation or verifying a sequence of computations, where each step can be proven individually and then combined. This approach could be beneficial in FL, where model aggregation often involves processing and verifying large, complex datasets. The application of recursive ZKPs in this context could enhance efficiency, making the overall process more manageable and scalable.

\smallskip \noindent\textbf{Power Consumption for Large-Scale Computing.}
Our results show that as the number of clients increases, both the synchronization time and aggregation time will experience an increase. This, in turn, also places a heightened computational burden on the centralized aggregator, too. It's worth noting that in practical scenarios, the centralized aggregator can be a sizable corporation (e.g., Google) equipped with the necessary resources to manage the computational costs and system power consumption efficiently. In this work, system power consumption is beyond the scope of discussion as a FL study. However, system power consumption is a non-trivial topic for system study and shall be discussed in future work, especially in large-scale computing setups.

\section{Conclusion} \label{sec: conclusion}
We present a novel and pioneering FL approach for the era of big data, \zkFL, which utilizes \ZKPs to ensure a trustworthy aggregation process on the centralized aggregator. Through rigorous theoretical analysis, we establish that \zkFL effectively addresses the challenge posed by a malicious aggregator during the model aggregation phase. Moreover, we extend \zkFL to a blockchain-based system, significantly reducing the verification burden on the clients. The empirical analysis demonstrates that our design achieves superior levels of security and privacy compared to traditional \FL systems, while maintaining a favorable training speed for the clients. These results showcase the practical feasibility and potential advantages of \zkFL and its blockchain-based version in real-world applications.

% \section*{Acknowledgment}
% The authors would like to thank Shuoying Zhang from FLock.io for the helpful discussion and industrial insights, and Lianfeng Zhou and Xinyang Wang from FLock.io for the experimental evaluation.

\bibliographystyle{ieeetr}
\bibliography{references}

\end{document}